\documentclass[%
 reprint,
 amsmath,amssymb,
 aps,
prstper,
]{revtex4-2}
\usepackage{subcaption}
\usepackage{graphicx}
\usepackage{dcolumn}
\usepackage{bm}
\usepackage[dvipsnames]{xcolor}
\usepackage[colorlinks=true, allcolors=blue]{hyperref}
\usepackage[nameinlink]{cleveref} 
\usepackage[most]{tcolorbox}
\usepackage[utf8]{inputenc}
\usepackage[T1]{fontenc}
\usepackage{booktabs}
\definecolor{softblue}{RGB}{245,247,250}
\definecolor{borderblue}{RGB}{210,220,230}
\definecolor{correctgreen}{RGB}{40,167,69}
\definecolor{wrongred}{RGB}{220,53,69}

\definecolor{prob90}{RGB}{204,235,204}  
\definecolor{prob70}{RGB}{214,238,214}  
\definecolor{prob60}{RGB}{219,240,219}  
\definecolor{prob10}{RGB}{242,248,242}  

\newtcolorbox{examplebox}[1][]{%
    enhanced,
    colback=white,  
    colframe=black,  
    boxrule=0.5pt,
    before skip=4pt,
    after skip=4pt,
    width=\columnwidth,
    breakable,
    title={\footnotesize\bfseries\itshape Example},
    #1
}
\hypersetup{
    colorlinks=true,       
    linkcolor=MidnightBlue, 
    citecolor=ForestGreen, 
    urlcolor=RoyalBlue,    
    filecolor=Purple,      
    menucolor=Gray,        
}

\begin{document}

\title{On the Reasoning Capacity of AI Models and How to Quantify It}

\author{\href{https://www.linkedin.com/in/santoshkumarradha/}{\hspace{1mm}Santosh Kumar Radha} }
\altaffiliation[Also at ]{Agnostiq Inc.}
\email{santosh@agnostiq.ai}
\affiliation{%
325 Front St W\\
Toronto, ON M5V 2Y1\\
Agnostiq Inc.\\
}%

\author{\href{https://www.linkedin.com/in/oktay-goktas-8b2881167/}{\hspace{1mm}Oktay Goktas}}
\email{oktay@agnostiq.ai}
\affiliation{%
325 Front St W\\
Toronto, ON M5V 2Y1\\
Agnostiq Inc.\\
}%


\begin{abstract}
    Recent advances in Large Language Models (LLMs) have intensified the debate surrounding the fundamental nature of their reasoning capabilities. While achieving high performance on benchmarks such as GPQA and MMLU, these models exhibit limitations in more complex reasoning tasks, highlighting the need for more rigorous evaluation methodologies.  We propose a novel phenomenological approach that goes beyond traditional accuracy metrics to probe the underlying mechanisms of model behavior, establishing a framework that could broadly impact how we analyze and understand AI systems. Using positional bias in multiple-choice reasoning tasks as a case study, we demonstrate how systematic perturbations can reveal fundamental aspects of model decision-making. To analyze these behaviors, we develop two complementary phenomenological models: a Probabilistic Mixture Model (PMM) that decomposes model responses into reasoning, memorization, and guessing components and an Information-Theoretic Consistency (ITC) analysis that quantifies the relationship between model confidence and strategy selection. Through controlled experiments on reasoning benchmarks, we show that \textit{true reasoning} remains challenging for current models, with apparent success often relying on sophisticated combinations of memorization and pattern matching rather than genuine logical deduction. More fundamentally, we demonstrate that \textit{accuracy alone often overstates a model's reasoning abilities}, as model behavior can be characterized through underlying mechanisms in the phase space of cognitive strategies, revealing how models dynamically balance different approaches when responding to queries. This framework enables quantitative criteria for real-world deployments, allowing applications to specify reliability thresholds based on strategy distributions rather than aggregate performance metrics. By establishing principled methods for probing and quantifying reasoning behavior, our work provides both theoretical insights into model capabilities and practical tools for developing more reliable evaluation methodologies.
\end{abstract}
\maketitle
\onecolumngrid
\begin{figure}[!h]
    \centering
    \includegraphics[width=\textwidth]{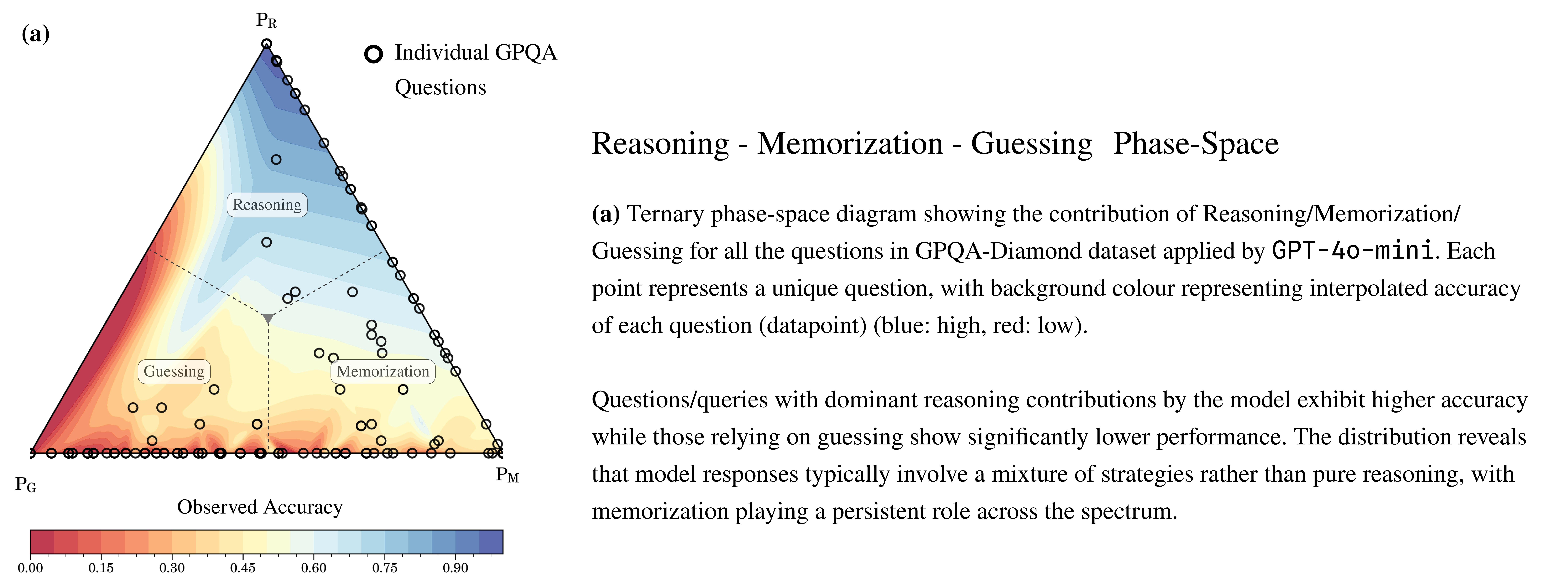}
    \label{fig:hero}
\end{figure}
\twocolumngrid

\onecolumngrid
\clearpage
\twocolumngrid

\newpage
\section{Introduction}

The emergence of increasingly capable Large Language Models (LLMs) has raised fundamental questions about the nature of artificial intelligence and its capacity for genuine reasoning. While models like GPT-4, Claude, LLama and Gemini demonstrate remarkable performance across wide spectrum of complex tasks  \cite{openai2023gpt,touvron2023llama,team2023gemini}, distinguishing between true logical deduction and non-deductive cognitive processes remains a central challenge in AI research  \cite{bubeck2023sparks,wei2022chain}. Traditional evaluation methodologies, primarily centered on benchmark performance and accuracy metrics, have provided valuable but incomplete insights into model capabilities, with standard reasoning benchmarks such as GSM8K  \cite{cobbe2021training}, GPQA  \cite{rein2023gpqa}, and Big-Bench  \cite{suzgun2022challenging} demonstrating impressive yet potentially misleading performance figures. Recent investigations systematically challenge these results - experiments with GSM-Symbolic reveal significant performance degradation under minor question reformulations despite preserved logical structure  \cite{mirzadeh2024gsm}, while analyses of other benchmarks demonstrate how success often stems from dataset-specific regularities rather than genuine logical inference  \cite{wan2024b,wu2023reasoning}. 
\begin{figure}[!htp]
    \centering
    \includegraphics[width=\linewidth]{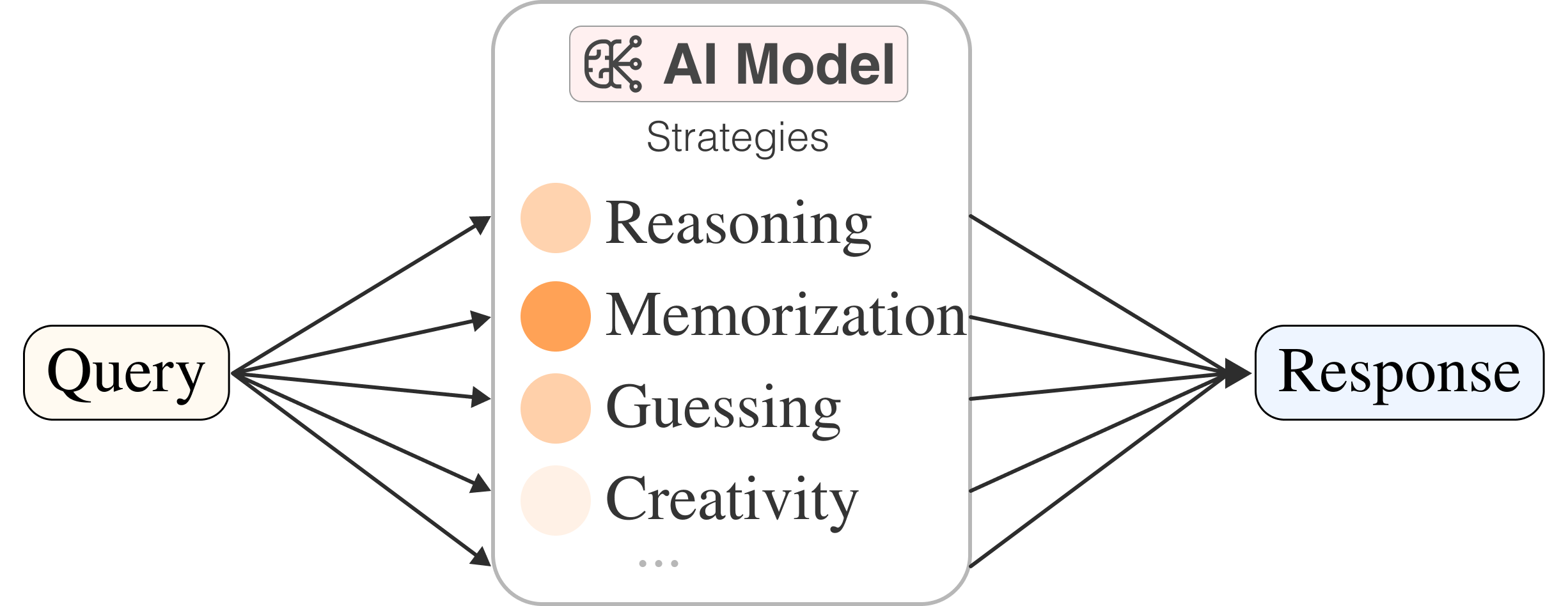}
    \caption{Phenomenological decomposition of language model behavior through strategy space, illustrating the mappings between input queries and output responses. }
    \label{fig:strategy-decomposition}
\end{figure}
These findings, coupled with the increasing deployment of LLMs in domains ranging from scientific research to educational assessment, highlight the limitations of current evaluation frameworks and underscore the need for comprehensible benchmarks and explainable AI (xAI) \cite{gunning2019darpa} approaches to elucidate the mechanisms underlying model behavior. 
Increasingly widespread usage of LLMs, and AI models in critical domains such as scientific research and educational assessment necessitates methodologies that can systematically characterize and interpret the fundamental processes driving model predictions \cite{chen2024survey}. The ability to distinguish between genuine reasoning and other cognitive processes becomes particularly crucial in applications where verifiable logical deduction directly impacts system reliability and user outcomes.

Existing approaches to understanding model reasoning capabilities have primarily focused on enhancing evaluation frameworks through techniques such as Chain-of-Thought prompting  \cite{wei2022chain}, Iteration-of-Thought  \cite{radha2024iteration}, and self-consistency checks  \cite{wang2022self}. While these methods demonstrate improved performance on reasoning benchmarks, they primarily operate by structuring the model's input and output format rather than providing insights into the underlying decision-making processes. Quantitative analyses of these enhanced frameworks reveal that improvements often arise from better exploitation of model priors rather than enhanced logical deduction capabilities  \cite{dziri2024faith,razeghi2022impact}. For instance, contemporary studies demonstrate that models can achieve high accuracy on reasoning tasks while producing logically inconsistent intermediate steps  \cite{jiang2024peek,gendron2023large}, suggesting that apparent reasoning success may emerge from sophisticated pattern matching rather than systematic logical inference.

The fundamental limitation of current evaluation methodologies lies in their treatment of model behavior as a black-box optimization problem, where success is primarily measured through aggregate performance metrics such as accuracy and F1 scores. While computationally tractable, this approach fails to capture the dynamic interplay between different cognitive strategies employed by these models. More critically, it provides limited insight into how models transition between different decision-making modes when confronted with varying task complexities or input perturbations. These limitations become particularly evident when examining how model performance varies systematically under controlled modifications to input structure, suggesting the need for evaluation frameworks that can probe and quantify the fundamental mechanisms underlying model behavior.

To address these methodological limitations, we propose adopting a phenomenological modeling approach to AI systems analysis. In physical scientific domains, phenomenological models $\mathcal{M}$ provide valuable insights by constructing theoretical frameworks of complex underlying systems that capture observable behaviors under specific experimental conditions. While phenomenological approaches have been successfully applied to understand neural network dynamics and training trajectories  \cite{tovey2023towards, mei2019mean}, their application to language models remains largely unexplored. While such models may not fully characterize the underlying system, they enable quantitative predictions within well-defined regimes and offer systematic approaches to understanding complex phenomena. We develop two such models: a Probabilistic Mixture Model (PMM) that decomposes cognitive strategies (\cref{fig:strategy-decomposition}) and an Information-Theoretic Consistency (ITC) analysis that quantifies decision-making uncertainty. Applied to language models, this framework enables the decomposition of model behavior into measurable components through controlled perturbations of input conditions. Consider a language model $\mathcal{B}$ operating on input query space $\mathcal{X}$. Traditional evaluation approaches treat this as a mapping $\mathcal{B}: \mathcal{X} \rightarrow \mathcal{Y}$, where $\mathcal{Y}$ represents the output response space. In contrast, our phenomenological approach introduces an intermediate representation $\mathcal{B}: \mathcal{X} \xrightarrow{\kappa} \mathcal{S} \xrightarrow{\omega} \mathcal{Y}$, where $\mathcal{S}$ represents a space of cognitive strategies, and mappings $\kappa$, $\omega$ characterize how the model selects and applies these strategies. This decomposition, while necessarily an approximation, provides testable predictions about model behavior under systematic perturbations.

The power of this approach lies not in providing a complete description of model internals but in establishing a quantitative framework for analyzing observable model behaviors. We can characterize how models transition between different cognitive modes and predict their behavior under specific conditions by introducing controlled experimental variables and measuring their effects on strategy selection and application. This framework enables the development of rigorous evaluation criteria based on mechanistic understanding rather than aggregate performance metrics, with specific predictions that can be experimentally verified  \cite{golgoon2024mechanistic,bereska2404mechanistic}.  While our investigation focuses on large language models and employs positional bias as a primary experimental probe, this methodological framework offers a systematic basis for quantifying model capabilities and behavioral characteristics across diverse machine learning architectures through rigorous empirical analysis.

To demonstrate the utility of our phenomenological framework, we examine positional bias in multiple-choice reasoning tasks as a controlled experimental probe. Positional effects in language models provide an ideal test case - they are systematically measurable, arise from fundamental architectural properties and critically influence model behavior. By introducing an order parameter $\theta \in [0,1]$ that controls position randomization, we systematically study how models transition between different cognitive strategies as positional information becomes increasingly unreliable. Our analysis reveals rich structure in the strategy probability space $\mathcal{S}$, including stable attractors that characterize how models balance reasoning and memorization and phase transitions that mark fundamental shifts in cognitive approach. The PMM decomposition demonstrates that true reasoning - characterized by position-invariant logical deduction - remains challenging for current models, with apparent success often relying on sophisticated memorization and pattern-matching combinations. Complementary ITC analysis reveals systematic relationships between strategy selection and prediction uncertainty, providing quantitative criteria for distinguishing between different cognitive modes. These findings extend beyond positional bias, offering a general framework for analyzing how language models combine different cognitive approaches when solving complex tasks. The identification of conserved quantities in strategy evolution and the characterization of phase transitions in cognitive space provide both theoretical insights into model capabilities and practical metrics for evaluating model reliability in deployment scenarios.

These findings extend beyond positional bias, offering a general framework for analyzing how language models combine different cognitive approaches when solving complex tasks. Integrating phenomenological modeling with controlled experimental probes enables systematic characterization of model behavior, providing both theoretical insights and practical evaluation criteria. This systematic approach to understanding language models through the lens of observable behaviors and controlled perturbations represents a significant advancement in AI systems analysis.

This paper makes several fundamental contributions to the study of artificial intelligence:
\begin{itemize}
    \item Introduction of a phenomenological framework for analyzing language models, establishing principled methods for characterizing model behavior through controlled experimental manipulation rather than treating models as black boxes
    \item Development of a probabilistic mixture model that decomposes model responses into fundamental cognitive strategies, providing quantitative metrics for distinguishing between genuine reasoning and other decision-making processes
    \item Establishment of information-theoretic bounds on model uncertainty across different cognitive modes, enabling rigorous evaluation of prediction confidence and strategy selection
\end{itemize}

The remainder of this paper is organized as follows. \Cref{sec:related-work} reviews related work in reasoning evaluation and positional bias. \Cref{sec:models-framework} introduces our models and theoretical framework. \Cref{sec:results} presents experimental results and analysis, discusses the implications, and concludes.

\section{Related Work: Benchmarks, Bias, and Reasoning in Language Models}\label{sec:related-work}
Reasoning benchmarks such as GPQA, GSM8K, and Big-Bench have become vital tools for evaluating the logical and deductive reasoning capabilities of LLMs across diverse tasks, including arithmetic, abstract problem-solving, and contextual understanding \cite{rein2023gpqa,cobbe2021training,suzgun2022challenging}. Techniques such as CoT prompting and its iterative extension, Iteration-of-Thought (IoT), Tree-of-Thought (ToT) have demonstrated notable improvements in reasoning performance by guiding models to produce intermediate reasoning steps, enabling more structured logical deduction and enhanced problem-solving  \cite{wei2022chain, radha2024iteration,yao2024tree}. However, recent studies have questioned whether these methods and benchmarks genuinely assess reasoning abilities or primarily capture the models' reliance on dataset-specific heuristics and statistical patterns  \cite{valmeekam2022large,dasgupta2022language}. Approaches like GSM-Symbolic have tackled these concerns by employing symbolic templates for controlled question generation, revealing substantial performance declines when faced with slight changes in phrasing or irrelevant numerical adjustments, even when the underlying reasoning process remains constant \cite{mirzadeh2024gsm}. These findings highlight the prevalence of surface-level memorization and heuristic reliance, emphasizing critical gaps in benchmark design. Beyond arithmetic, similar limitations are evident in broader reasoning tasks, where positional biases, token sensitivities, and structural cues often inflate performance metrics and obscure genuine reasoning capabilities \cite{jia2017adversarial,mccoy2019right}. For example, experiments with GSM-NoOp demonstrate that models falter when irrelevant details are introduced, causing notable performance drops  \cite{mirzadeh2024gsm}. These challenges underscore the need for dynamic evaluations that disentangle reasoning from heuristic exploitation, building on prior work that has laid a foundation for robust and nuanced reasoning assessments  \cite{dziri2024faith,razeghi2022impact,jiang2024peek,gendron2023large}.

Positional bias is a common phenomenon in modern language models, where the position of an input disproportionately affects the model's predictions, regardless of its logical or contextual relevance \cite{wang2024eliminating,pezeshkpour2023large}. This bias manifests across diverse tasks, including question answering, ranking, and reasoning, often undermining the robustness of model performance \cite{guo2024serial,zheng2023large}. Mechanistically, positional bias arises from components like causal attention and positional embeddings, which favor specific token positions in input sequences, leading to systematic disparities in predictions based on input order. Quantifying this bias, however, is inherently challenging, as metrics such as Standard Deviation of Recalls (RStd) and Relative Standard Deviation (RSD) provide only task-specific insights and fail to capture its nuanced effects in broader contexts like reasoning evaluation  \cite{choi2024mitigating,yu2024mitigate}. Prior works such as  \citep{choi2024mitigating} have predominantly sought to mitigate or eliminate positional bias, introducing techniques such as bidirectional attention mechanisms or reassigning token positions based on importance scores to improve performance \cite{huang2023bias}. Unlike prior efforts that treat positional bias as a flaw to mitigate, our work leverages it as a diagnostic tool to probe reasoning in LLMs. By isolating and manipulating positional effects within a reasoning benchmark, we evaluate whether model predictions arise from genuine logical deduction or reliance on probabilistic tendencies. This approach goes beyond static performance evaluations, using positional bias to reveal latent dependencies and assess the reasoning processes shaping model outputs. Our study reframes positional bias not as an error to eliminate but as a novel lens for understanding and dissecting reasoning capabilities in LLMs.

Prior attempts to develop phenomenological neural network models have focused primarily on training dynamics, signal propagation, and emergent behaviors. Mean-field theories have successfully characterized network behavior at initialization  \cite{mei2019mean}, particularly in analyzing quantization effects and depth-precision trade-offs  \cite{blumenfeld2019mean}. Recent extensions of these approaches to modern architectures demonstrate how theoretical frameworks can illuminate fundamental properties of neural systems, including phase transitions in learning dynamics  \cite{feng2021phases} and emergent behavioral patterns. However, applications of such systematic theoretical analysis to reasoning models remain sparse, with most works focusing on empirical performance metrics rather than underlying mechanistic properties. Our work bridges this gap by developing phenomenological models that capture observable behaviors under controlled perturbations, enabling quantitative predictions about strategy selection and decision-making processes in artificial reasoning.

\section{Models and Framework}\label{sec:models-framework}

The GPQA benchmark has emerged as a powerful tool for evaluating reasoning capabilities in language models, offering a diverse set of multiple-choice questions that probe various aspects of logical and analytical thinking. While positional bias is a well-documented phenomenon, its manifestation and impact can vary significantly depending on the specific model architecture, model parameters, and dataset characteristics. In this work, we first develop a systematic framework to analyze these position-dependent effects in the context of \texttt{GPT-4o-mini}'s performance on GPQA. This focused examination serves two purposes: first, it provides detailed insights into how positional information influences reasoning behavior in a specific model-benchmark pair, and second, it establishes a methodological foundation for broader investigations of our reasoning in the next section.

\subsection{Positional Bias Analysis}
The core of our analysis centers on the GPQA benchmark, a multiple-choice reasoning evaluation framework where each question presents four options, exactly one of which is correct. Let $\mathcal{Q} = \{q_n \mid n \in [1, M]\}$ denote our question set, where $M$ is the total number of questions. Each question $q \in \mathcal{Q}$ is associated with one correct answer $c(q)$ and three incorrect options $\{w_1(q), w_2(q), w_3(q)\}$. These options are presented in fixed positions drawn from the set $\mathcal{O} = \{A, B, C, D\}$.

We treat the language model $\mathcal{B}$ as a black-box system, which for any given question $q$ and its associated options, generates a probability distribution over the possible answer positions:
\begin{equation}
    \mathcal{B}(q) \mapsto \mathbf{\Pi}(q) = [\pi_A, \pi_B, \pi_C, \pi_D]
    \label{eq:prob_dist}
\end{equation}
where $\pi_o \in [0, 1]$ represents the probability assigned to position $o \in \mathcal{O}$, and $\sum_{o \in \mathcal{O}} \pi_o = 1$.

\begin{examplebox}
For a question "What is $2 + 2$?", a model might output probabilities $\mathbf{\Pi}(q) = [0.7, 0.2, 0.05, 0.05]$, indicating 70\% confidence in position A containing the correct answer. This distribution directly reveals position-dependent decision making.
\end{examplebox}

To systematically evaluate positional dependencies, we introduce an order parameter $\theta \in [0, 1]$ that determines the degree of position randomization in our experiments. At $\theta = 0$ correct answers maintain fixed positions across all questions, while positions are fully randomized at $\theta = 1$.  For intermediate values, $\theta$ represents the fraction of questions subject to position randomization.

\begin{examplebox}
onsider two questions ``What is 2 + 2?'' and ``What is 3 + 3?'':

At $\theta = 0$:
Correct answers fixed in position A
\begin{itemize}
    \item Q1 always presented as: A:4, B:3, C:5, D:6
    \item Q2 always presented as: A:6, B:7, C:5, D:8
\end{itemize}

At $\theta = 0.5$:
Half the presentations maintain fixed positions, half use randomized positions like
\begin{itemize}
        \item Q1: A:6, B:4, C:3, D:5 (Randomized)
        \item Q2: A:6, B:7, C:5, D:8 (Fixed)
\end{itemize}

\end{examplebox}

The position-specific accuracy $\alpha_o(q,\theta)$ for each position $o \in \mathcal{O}$ is defined as:
\begin{equation}
    \alpha_o(q,\theta) = \frac{1}{N(\theta)} \sum_{n=1}^{N(\theta)} \mathbb{I}[\mathcal{B}(q_n) = o \mid c(q_n) = o]
    \label{eq:pos_acc}
\end{equation}
where $N(\theta)$ represents the number of trials at a given $\theta$ value, and $\mathbb{I}[\cdot]$ is the indicator function. 

\begin{examplebox}
To illustrate $\alpha_o(q,\theta)$, consider our "2 + 2" question across different randomization levels with correct answer placed at position $o$:

At $\theta = 0$ (fixed positions):
\begin{align*}
    \alpha_A(q,0) &= 0.8 \text{ (memorized position)}\\
    \alpha_B(q,0) &= 0.4\\
    \alpha_C(q,0) &= 0.4\\
    \alpha_D(q,0) &= 0.3
\end{align*}

At $\theta = 0.5$ (partial randomization):
\begin{align*}
    \alpha_A(q,0.5) &= 0.6\\
    \alpha_B(q,0.5) &= 0.5\\
    \alpha_C(q,0.5) &= 0.5\\
    \alpha_D(q,0.5) &= 0.4
\end{align*}

At $\theta = 1$ (full randomization):
\begin{align*}
    \alpha_o(q,1) &\approx 0.3 \text{ for all positions $o$}
\end{align*}

Position-specific accuracies would converge as randomization increases, leading to the model's true performance. 
\end{examplebox}

To characterize question difficulty and positional dependence, we introduce two complementary metrics. The position-averaged accuracy $\mu(q)$ captures the model's overall performance on a question:
\begin{equation}
    \mu(q,\theta) = \frac{1}{|\mathcal{O}|} \sum_{o \in \mathcal{O}} \alpha_o(q,\theta)
    \label{eq:avg_acc}
\end{equation}
while the position-dependent variance $\sigma^2(q)$ quantifies the variation in performance across positions:
\begin{equation}
    \sigma^2(q,\theta) = \frac{1}{|\mathcal{O}|} \sum_{o \in \mathcal{O}} (\alpha_o(q,\theta) - \mu(q,\theta))^2
    \label{eq:pos_var}
\end{equation}

These metrics combine to form a two-dimensional difficulty map $\Psi(q,\theta) = ( \mu(q,\theta), \sigma^2(q,\theta))$, which proves instrumental in categorizing questions based on their difficulty and susceptibility to positional bias. Questions with high $\mu$ and low $\sigma^2$ indicate consistent reasoning ability, while high $\sigma^2$ values, regardless of $\mu$, suggest strong positional dependencies that may undermine genuine reasoning assessment.

\subsection{Probabilistic Mixture Model}

To dissect model behavior, we develop a framework decomposing responses into three fundamental cognitive strategies: memorization, reasoning, and guessing. This decomposition, while not unique, provides a natural basis for quantifying the balance between pattern matching and logical inference in positional reasoning tasks. As we will see in \cref{subsec:result-pmm}, the model's empirical validation and observed deviations suggest potential extensions for different experimental conditions while confirming its utility for the studied tasks.

To begin with, building on our earlier notation, we consider the black box model $\mathcal{B}$ operating on questions $q \in \mathcal{Q}$ with options in positions $\mathcal{O}$. For each question, we hypothesize that the model employs a mixture of strategies with probabilities $P_M(q)$, $P_R(q)$, and $P_G(q)$, representing the probability of using memorization, reasoning, and guessing strategies respectively for question $q$. These probabilities sum to one (complete basis assumption):
\begin{equation}
    P_M(q) + P_R(q) + P_G(q) = 1
    \label{eq:strategy_norm}
\end{equation}

The probability of a correct response under this mixture model for a question $q$ and position $o \in \mathcal{O}$ is:
\begin{align}
    P_{\text{correct}}(q, o) &= P_M(q) \cdot P_{\text{M}}(o) \nonumber \\
    &\quad + P_R(q) \cdot P_{\text{R}} \nonumber \\
    &\quad + P_G(q) \cdot P_{\text{G}}
    \label{eq:p_correct_mix}
\end{align}
where $P_{\text{M}}(o)$ represents the success probability under memorization strategy for position $o$, while $P_{\text{R}}$ and $P_{\text{G}}$ are position-independent success probabilities for reasoning and guessing strategies, respectively. Intuitively, this equation describes how the model combines three different approaches to answer a question: when it uses memorization ($P_M(q)$), its success depends on the specific position; when it uses reasoning ($P_R(q)$), it has a fixed success rate regardless of position; and when it guesses ($P_G(q)$), it has the baseline random chance of success.
\begin{examplebox}
Consider a model evaluating "What is the capital of France?" with strategy probabilities $P_M(q) = 0.3$, $P_R(q) = 0.6$, $P_G(q) = 0.1$. With perfect reasoning ($P_R = 1$) and random guessing ($P_G = 0.25$), \cref{eq:p_correct_mix} yields:
\[P_{\text{correct}}(q,o) = 0.3P_M(o) + 0.6 + 0.025\]
This decomposition shows 60\% of success probability comes from reasoning, while position-dependent memorization contributes 30\%.
\end{examplebox}

For the memorization strategy, we model position dependence as:
\begin{equation}
    P_{\text{M}}(o) = 
    \begin{cases}
        1, & \text{if } o = o_m \\
        P_{\mathcal{O}}, & \text{if } o \neq o_m
    \end{cases}
    \label{eq:p_mem}
\end{equation}
where $o_m$ is the memorized position. The base probability for random selection, $P_{\mathcal{O}} = 1/|\mathcal{O}|$, represents the pure guessing probability in the absence of reasoning or memorization. Since reasoning and memorization probabilities are explicitly accounted for, $1/|\mathcal{O}|$ represents the pure guessing probability, isolating the uniform contribution of random selection.
\begin{examplebox}
To illustrate the memorization probability function defined in \cref{eq:p_mem}, consider our previous quesion "What is 2 + 2?" with options presented across positions $\mathcal{O}$. Let us examine a model that has developed a strong positional association, memorizing position A ($o_m = A$) as the likely location for correct answers. According to \cref{eq:p_mem}, when the correct answer "4" appears in position A, the memorization contribution $P_{\text{M}}(o)$ equals 1, reflecting complete positional dependence. However, when "4" appears in any other position ($o \neq o_m$), the model defaults to uniform random selection with probability $P_{\mathcal{O}} = 1/|\mathcal{O}| = 0.25$. This behavior manifests as:

\begin{equation}
P_{\text{M}}(o) = 
\begin{cases}
1, & \text{if } o = A \\
0.25, & \text{if } o \in \{B, C, D\}
\end{cases}
\end{equation}

This formulation quantifies how positional memorization can significantly bias model responses, potentially masking true reasoning capabilities.
\end{examplebox}
Moving forward, we consider ideal conditions where if the model employs reasoning, it does so perfectly ($P_{\text{R}} = 1$), and if it guesses, it does so randomly ($P_{\text{G}} = P_{\mathcal{O}}$). The assumption of perfect reasoning ($P_{\text{R}} = 1$) establishes a theoretical baseline where reasoning, if applied, would always yield the correct answer no matter where the position of correct answer is. This doesn't imply that real-world reasoning is perfect—it creates a benchmark for analyzing how much the model's actual behavior deviates from ideal reasoning due to factors like computational errors or incomplete understanding. Under ideal conditions ($P_{\text{R}} = 1$ and $P_{\text{G}} = P_{\mathcal{O}}$), the probability of correct response becomes: 
\begin{align}
    P_{\text{correct}}(q, o) = 
    \begin{cases}
        \begin{aligned}[t]
            P_M(q) & + P_R(q) \\
            & + P_{\mathcal{O}} P_G(q),
        \end{aligned}
        & \text{if } o = o_m \\
        \begin{aligned}[t]
            P_{\mathcal{O}} P_M(q) & + P_R(q) \\
            & + P_{\mathcal{O}} P_G(q),
        \end{aligned}
        & \text{if } o \neq o_m
    \end{cases}
    \label{eq:p_correct_cases}
\end{align}

To estimate these probabilities from empirical data, we observe accuracies in two conditions: 
\begin{align}
    A_{o_m}(q) &= \frac{\text{Correct predictions when } o = o_m}{\text{Total trials when } o = o_m} \label{eq:acc_mem} \\
    A_{\text{other}}(q) &= \frac{\text{Correct predictions when } o \neq o_m}{\text{Total trials when } o \neq o_m} \label{eq:acc_other}
\end{align}

\begin{examplebox}
Suppose we evaluate our model on 100 trials of the question, with position A as the memorized position ($o_m = A$). If the model correctly answers 80 trials when the answer is in position A, and 45 trials when the answer is in other positions, we obtain:
\begin{align}
A_A(q) &= \frac{80}{100} = 0.8 \\
A_{\text{other}}(q) &= \frac{45}{100} = 0.45
\end{align}
This substantial accuracy differential ($\Delta A(q) = 0.35$) would suggest significant positional bias in the model's decision-making process.
\end{examplebox}

These empirical accuracies directly correspond to the probabilities in \cref{eq:p_correct_cases}:
\begin{align}
    A_{o_m}(q) &= P_M(q) + P_R(q) + P_{\mathcal{O}} P_G(q) \label{eq:acc_mem_model} \\
    A_{\text{other}}(q) &= P_{\mathcal{O}} P_M(q) + P_R(q) + P_{\mathcal{O}} P_G(q) \label{eq:acc_other_model}
\end{align}

Taking the difference eliminates terms that don't depend on position:
\begin{align}
    \Delta A(q) &= A_{o_m}(q) - A_{\text{other}}(q) \nonumber \\
    &= \big[ P_M(q) + P_R(q) + P_{\mathcal{O}} P_G(q) \big] \nonumber \\
    &\quad - \big[ P_{\mathcal{O}} P_M(q) + P_R(q) + P_{\mathcal{O}} P_G(q) \big] \nonumber \\
    &= P_M(q) - P_{\mathcal{O}} P_M(q) \nonumber \\
    &= P_M(q)(1 - P_{\mathcal{O}})
    \label{eq:delta_acc}
\end{align}

This key equation isolates the memorization probability:
\begin{equation}
    P_M(q) = \frac{\Delta A(q)}{1 - P_{\mathcal{O}}}
    \label{eq:p_mem_solve}
\end{equation}
This equation reveals how strongly the model relies on memorization by measuring the difference in performance between memorized and non-memorized positions. A larger difference ($\Delta A(q)$) indicates stronger memorization behavior, while a smaller difference suggests more position-independent reasoning or guessing.

Substituting this back into \cref{eq:acc_mem_model} gives:
\begin{align}
\begin{split}
    A_{o_m}(q) &= P_M(q) + P_R(q) + P_{\mathcal{O}}[1 - P_M(q) - P_R(q)] \nonumber \\
    &= [P_M(q) + P_R(q)](1 - P_{\mathcal{O}}) + P_{\mathcal{O}}
    \label{eq:acc_mem_expand}
\end{split}
\end{align}

Solving for the combined probability:
\begin{equation}
    P_M(q) + P_R(q) = \frac{A_{o_m}(q) - P_{\mathcal{O}}}{1 - P_{\mathcal{O}}}
    \label{eq:p_combined}
\end{equation}

With $P_M(q)$ known from \cref{eq:p_mem_solve}, we can determine:
\begin{align}
    P_R(q) &= \frac{A_{o_m}(q) - P_{\mathcal{O}}}{1 - P_{\mathcal{O}}} - P_M(q) \label{eq:p_reason} \\
    P_G(q) &= 1 - P_M(q) - P_R(q) \label{eq:p_guess}
\end{align}

\begin{examplebox}
Following our previous geography question example with $A_A(q) = 0.8$ and $A_{\text{other}}(q) = 0.45$, we can decompose the model's strategy probabilities. Using \cref{eq:p_mem_solve,eq:p_reason,eq:p_guess} with $P_{\mathcal{O}} = 0.25$:
\begin{align}
P_M(q) &= \frac{0.35}{0.75} \approx 0.47 \\
P_R(q) &= \frac{0.8 - 0.25}{0.75} - 0.47 \approx 0.26 \\
P_G(q) &= 1 - 0.47 - 0.26 = 0.27
\end{align}
This analysis reveals that the model relies predominantly on memorization (47\%), with reasoning and guessing contributing roughly equally (26\% and 27\% respectively) to its decision-making process.
\end{examplebox}

This systematic decomposition reveals how frequently the model employs each strategy, providing quantitative insights into its decision-making process. When combined with our positional bias metrics $\mu(q)$ and $\sigma^2(q)$ from \cref{eq:avg_acc,eq:pos_var}, this framework enables comprehensive analysis of model behavior across different question types and difficulty levels.

\subsection{Information-Theoretic Consistency Analysis}

While our probabilistic mixture model reveals the strategies employed by the language model, we require a complementary framework to assess the quality and calibration of its predictions. Our analysis involves several interrelated probability distributions: the model's raw output probabilities $\pi_o(q)$ (\cref{eq:prob_dist}), the accuracy-dependent reference probabilities we'll define below, and the empirical accuracy distribution. By examining the relationship between prediction uncertainty and accuracy through an information-theoretic lens, we can evaluate how well the model balances confidence with correctness.

Let $k = |\mathcal{O}|$ denote the number of options in our multiple-choice setup. For any question $q \in \mathcal{Q}$, we measure the uncertainty in the model's predictions using the Shannon entropy of its probability distribution over the option space:
\begin{equation}
    H(q) = -\sum_{i=1}^k \rho_i(q) \log_2 \rho_i(q)
    \label{eq:entropy}
\end{equation}
where $\rho_i(q)$ represents the probability of selecting option $i$ when it contains the correct answer. Lower entropy values indicate more concentrated probability mass (higher \textit{confidence)}, while higher values indicate more dispersed probabilities (higher \textit{uncertainty)}.

\begin{figure}
    \centering
    \includegraphics[width=\linewidth]{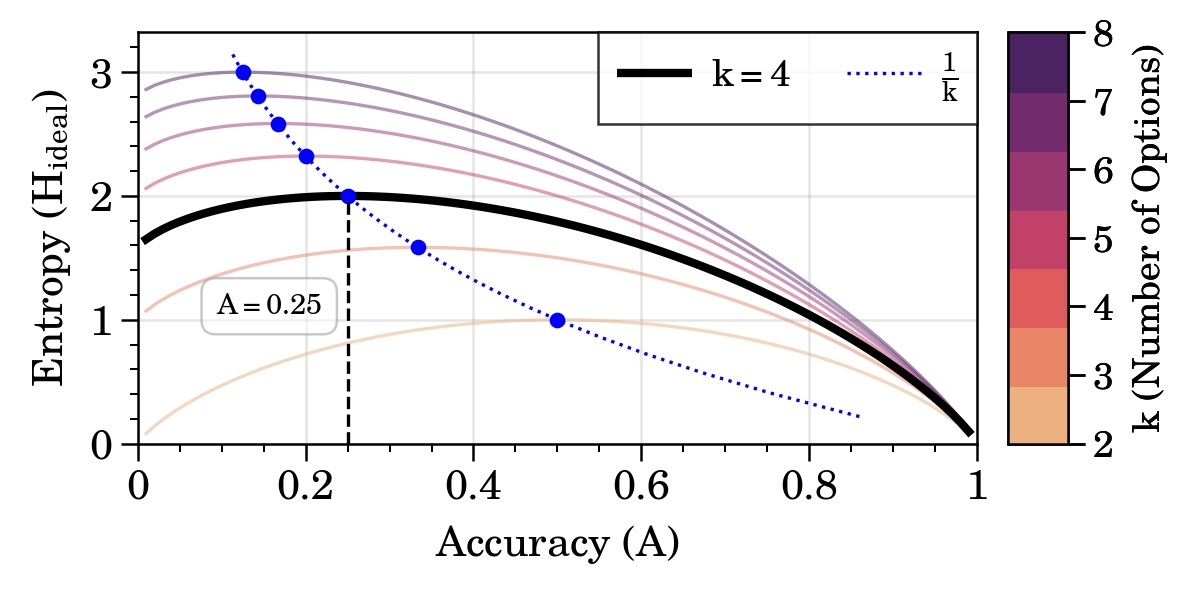}
    \caption{Theoretical entropy-accuracy frontiers for multiple-choice questions with varying numbers of options ($k$). The black curve represents the four-option case ($k=4$ where questions have multiple choices such A,B,C,D) relevant to GPQA, where maximum entropy ($H_{\text{max}} = 2$ bits) occurs at random guessing accuracy ($A = \frac{1}{k}=0.25$). The dotted blue line shows the relationship $A = \frac{1}{k}$, intersecting each frontier at its maximum entropy point. Higher values of $k$ result in larger maximum entropy due to increased uncertainty in the option space. The frontiers demonstrate how prediction uncertainty (entropy) varies with accuracy, reaching zero at perfect accuracy ($A = 1$) for all $k$.}
    \label{fig:entropy_frontier}
\end{figure}

To establish a theoretical baseline for analysis, we consider an ideal probabilistic model $(\mathcal{B}^*)$ that assigns probability $c$ to the correct option and distributes the remaining probability $(1-c)$ uniformly among the $k-1$ incorrect options:
\begin{equation}
    \rho_i(q) = 
    \begin{cases}
        c, & \text{if } i \text{ contains correct answer} \\
        \frac{1-c}{k-1}, & \text{otherwise}
    \end{cases}
    \label{eq:ideal_dist}
\end{equation}

The entropy of this idealized distribution defines our calibration frontier:
\begin{equation}
    H_{\text{ideal}}(c) = -c \log_2 c - (1-c) \log_2 \left(\frac{1-c}{k-1}\right)
    \label{eq:ideal_entropy}
\end{equation}
As shown in \cref{fig:entropy_frontier}, this frontier exhibits distinct characteristics across different values of $k$. The frontier represents a reference curve where probability mass is optimally distributed between correct and incorrect options, given a particular accuracy level. The behavior of $H_{\text{ideal}}(c)$ requires careful consideration at three critical points. Taking the limit as $c$ approaches these points:

At random guessing:
\begin{equation}
    H_{\text{max}}=\lim_{c \to 1/k}H_{\text{ideal}}(c) = -k \cdot \frac{1}{k} \log_2 \frac{1}{k} = \log_2 k
    \label{eq:max_entropy}
\end{equation}
reflecting maximum uncertainty when probabilities are uniform across all options.

At perfect prediction:
\begin{equation}
    H_{\text{min}}=\lim_{c \to 1}H_{\text{ideal}}(c) = 0 \text{ bits}
    \label{eq:min_entropy}
\end{equation}
indicating complete certainty in the prediction.

At complete inaccuracy:
\begin{equation}
    \lim_{c \to 0} H_{\text{ideal}}(c) = \log_2(k-1)
    \label{eq:limit_entropy}
\end{equation}
representing the entropy when probability is distributed uniformly across incorrect options.
These theoretical bounds become particularly illuminating for our specific analysis of GPQA with $k=4$ options. The maximum entropy at random guessing is $H_{\text{max}} = \log_2(4) = 2$ bits, while the limiting entropy as accuracy approaches zero is $H_{\text{ideal}}(c\rightarrow 0) = \log_2(3) \approx 1.58$ bits (\cref{fig:entropy_frontier}). This creates an asymmetric entropy profile, where complete uncertainty \textit{(random guessing)} produces higher entropy than systematic incorrectness. This asymmetry arises because random guessing represents complete disorder across all options, while zero accuracy implies a structured avoidance of the correct answer with uniform distribution across incorrect possibilities.

Since model accuracy $A(q)$ directly corresponds to the probability assigned to the correct option ($A(q) = c$), we can parameterize the entropy-accuracy frontier as:
\begin{equation}
    H_{\text{ideal}}(A) = -A \log_2 A - (1-A) \log_2 \left(\frac{1-A}{k-1}\right)
    \label{eq:entropy_frontier}
\end{equation}

The relationship between empirical entropy-accuracy pairs and this theoretical frontier provides deep insights into model calibration. Models operating near the frontier demonstrates how probabilities should be distributed between correct and incorrect answers at different accuracy levels. Deviations from this frontier characterize distinct types of miscalibration: Systematic deviations below the frontier indicate overconfidence, where models exhibit lower uncertainty than their accuracy warrants, potentially reflecting excessive reliance on memorized patterns or positional heuristics. Conversely, points consistently above the frontier suggest underconfidence, where models maintain unnecessarily high uncertainty despite reliable performance. When combined with the positional analysis metrics $\mu(q)$ and $\sigma^2(q)$ from \cref{eq:avg_acc,eq:pos_var}, this framework enables comprehensive evaluation of how positional dependencies influence both strategy selection and prediction uncertainty.

\section{Results and Discussion} \label{sec:results}
To systematically investigate positional bias in language model reasoning, we conducted experiments using the \texttt{GPT-4o-mini} model on the GPQA-Diamond dataset, comprising 198 curated questions spanning physics, mathematics, and computer science. Our experimental framework centers on manipulating the position of correct answers while maintaining question content, allowing us to isolate and quantify position-dependent effects on model behavior.

\begin{figure}[!htb]
    \centering
    \includegraphics[width=\linewidth]{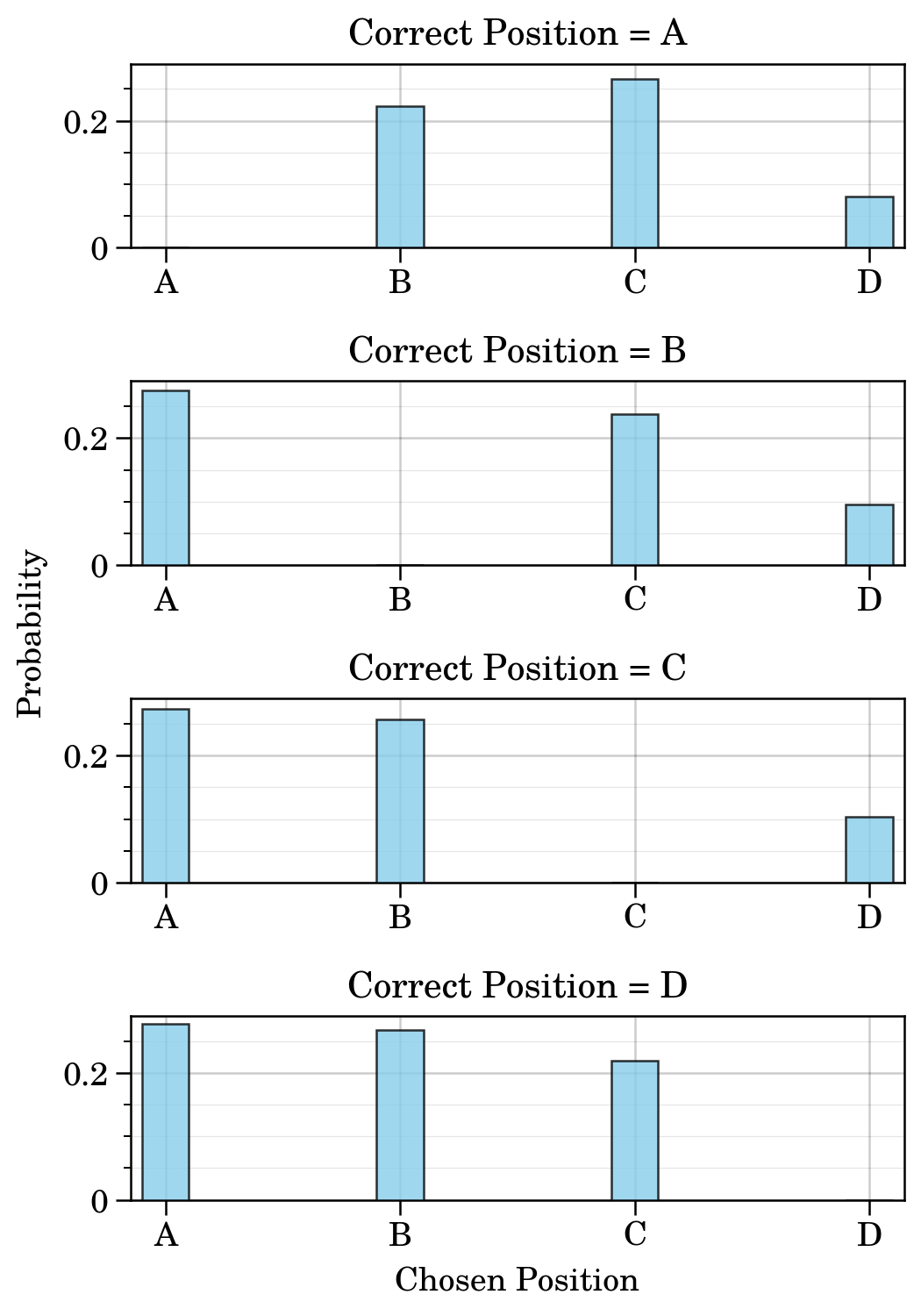}
    \caption{Wrong answer distributions conditional on correct answer position. Each subplot shows the probability mass of incorrect choices when the correct answer was at positions A, B, C, or D respectively. Position D exhibits markedly lower selection probability (~0.1) as an incorrect choice compared to other positions (~0.2-0.25), suggesting systematic positional bias. Probability mass is calculated as frequency of selection normalized by total number of trials.}
    \label{fig:wrong_answer_patterns}
\end{figure}
\begin{figure*}[!ht]
\centering
\begin{subfigure}[t]{0.48\textwidth}
\centering
\vspace{0pt}
\includegraphics[width=\linewidth]{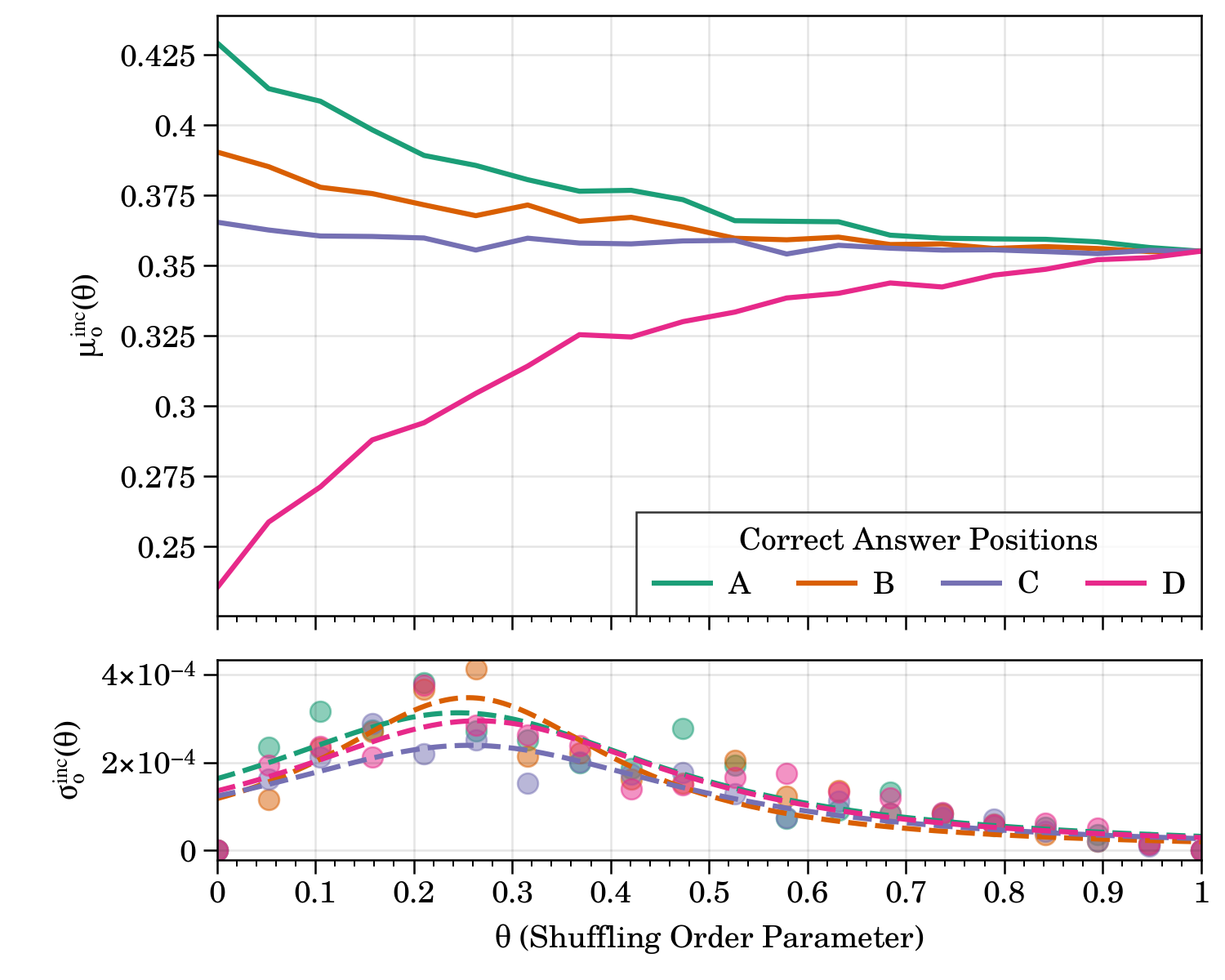}
\caption{Position-dependent accuracies ($\mu_o^{\text{inc}}(\theta)$) and variances ($\sigma_o^{\text{inc}}(\theta)^2$) under permissive randomization. he systematic degradation of performance reveals inherent positional dependencies, with distinct characteristics across answer positions that persist until complete randomization forces convergence to chance performance.}
\label{fig:acc_inc}
\end{subfigure}
\hfill
\begin{subfigure}[t]{0.48\textwidth}
\centering
\vspace{0pt}
\includegraphics[width=\linewidth]{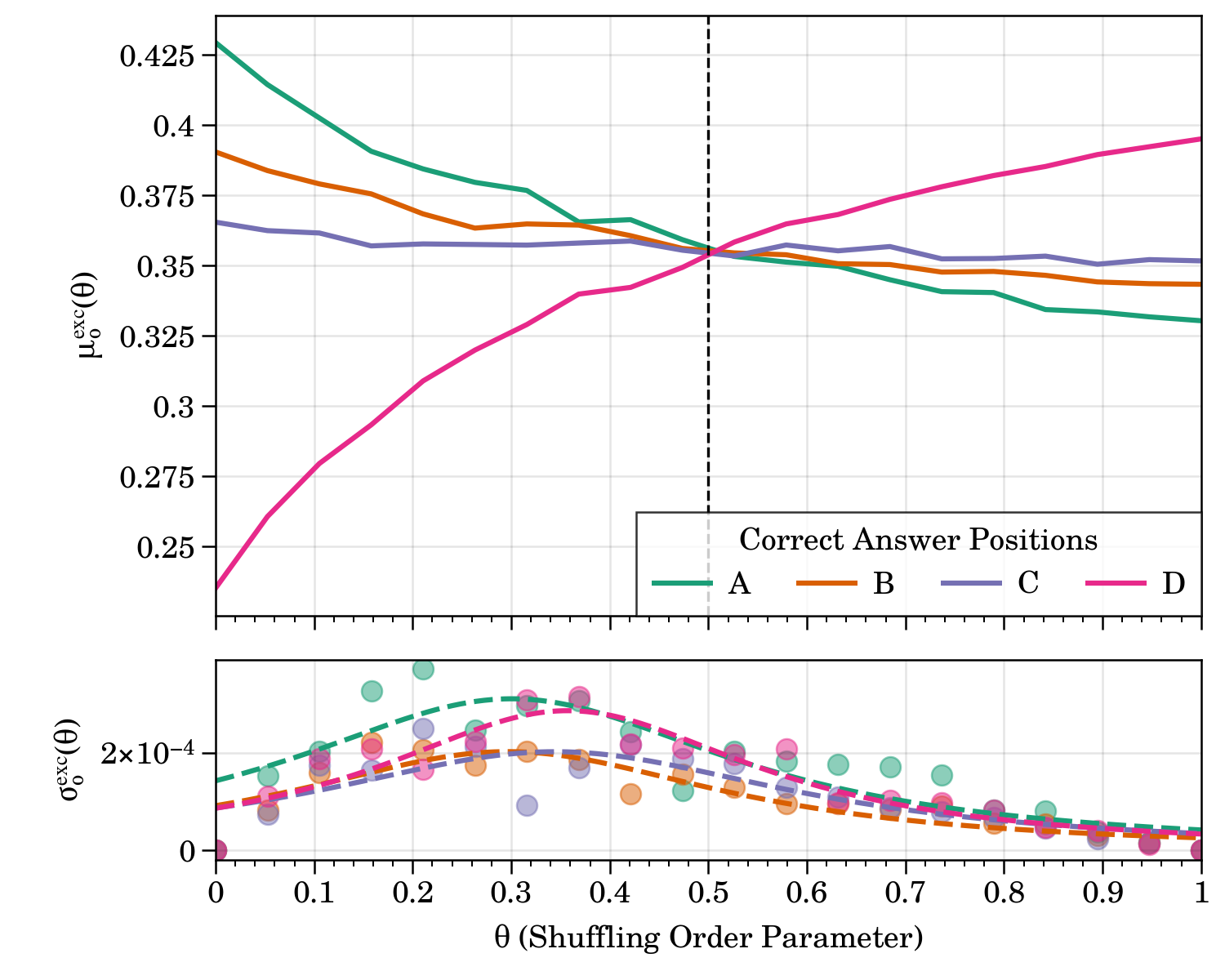}
\caption{Accuracies ($\mu_o^{\text{exc}}(\theta)$) and variances ($\sigma_o^{\text{exc}}(\theta)^2$) under constrained randomization where correct answers are excluded from original positions. The dynamics reveal a critical equilibrium at $\theta = 0.5$ where position-specific biases and randomization effects achieve perfect balance.}
\label{fig:acc_exc}
\end{subfigure}
\caption{Evolution of position-dependent model behavior under controlled randomization protocols. Upper panels track mean accuracies as functions of randomization parameter $\theta$, while lower panels show corresponding variance dynamics. The protocols differ fundamentally: inclusive randomization (a) permits correct answers at any position, while exclusive randomization (b) systematically prevents correct answers at original positions, enabling isolation of position-specific effects. This controlled comparison reveals both universal features (convergence to random chance) and protocol-specific dynamics (critical points, variance patterns) in the model's position-dependent behavior.}
\label{fig:combined_randomization}
\end{figure*}

Building on our formalism from \cref{eq:pos_acc}, we systematically vary the randomization parameter $\theta \in [0,1]$ to control the degree of position randomization, where $\theta = 0$ maintains original positions and $\theta = 1$ represents complete randomization. For each question $q \in \mathcal{Q}$, we generated 100 trials (repeated evaluations of the same question with different answer position configurations) where the correct answer was systematically placed at each position $o \in \mathcal{O} = \{A, B, C, D\}$, with incorrect options randomly distributed among remaining positions. This approach yields a comprehensive dataset of model responses across all possible answer configurations, enabling detailed analysis of positional dependencies. Crucially, we employed direct answer elicitation without additional reasoning prompts, such as Chain-of-Thought (CoT) or Iteration-of-Thought (IoT), focusing on the model's immediate response patterns. Exploring the impact of such reasoning-enhancing techniques remains an avenue for future investigation.

Our initial analysis examines the distribution of incorrect choices conditioned on the correct answer's position. For each correct position $o_c \in \mathcal{O}$, we calculate the probability of the model selecting an incorrect position $o_w$ as:
\begin{equation}
    \pi(o_w \mid o_c) = \frac{N(o_w \mid o_c)}{|Q|}
    \label{eq:wrong_choice_prob}
\end{equation}
where $N(o_w \mid o_c)$ represents the count of trials where position $o_w$ was chosen when the correct answer was at position $o_c$, and $|Q|$ is the total number of questions in our dataset. This analysis complements our position-specific accuracy metric $\alpha_o(q,\theta)$ by revealing how the model distributes its incorrect predictions, potentially highlighting systematic biases in its decision-making process.

Analysis of wrong answer distributions (\cref{fig:wrong_answer_patterns}) as a function of correct answer position reveals striking asymmetries in the model's position-dependent behavior. Most notably, position D exhibits markedly different characteristics across all distributions compared to positions A, B, and C, with consistently lower selection probabilities ($\pi(D \mid o_c) < 0.15$) when it represents an incorrect choice. This contrasts sharply with the relatively uniform selection probabilities ($\approx 0.2$-$0.25$) observed for other positions. Additionally, the data suggests a subtle preference for adjacent positions in incorrect selections, particularly when the correct answer occupies positions A or B. This spatial correlation effect diminishes as the correct answer approaches position D, potentially indicating an interaction between the model's positional encoding mechanism and its decision-making process. These patterns collectively suggest that the model's reasoning process and choice are significantly influenced by learned positional heuristics rather than purely logical deduction.

\subsection{Position-Dependent Accuracy Under Randomization as order-parameter}\label{subsec:position analysis}

Before delving into our more sophisticated modeling frameworks, we first establish foundational insights through a systematic analysis of positional effects under controlled randomization conditions. Building on our positional bias framework from \cref{eq:pos_acc}, we investigate two distinct randomization scenarios to probe the model's reliance on positional cues.  For each position $o \in \mathcal{O}$ and randomization parameter $\theta \in [0,1]$, our experimental protocol is as follows: We take a fraction $\theta$ of the questions and randomize their correct answer placements according to two distinct schemes, while for the remaining $(1-\theta)$ fraction, the correct answer remains at position $o$. In the first scheme (inclusive randomization), the correct answer can be placed at any position including $o$, while in the second (exclusive randomization), the correct answer is explicitly prevented from appearing at position $o$. This controlled approach enables us to disentangle different aspects of positional dependence.

We investigate position-dependent model behavior through two randomization protocols, systematically varying the placement of correct answers across positions. For inclusive randomization, we maintain the original position-specific accuracy metric $\alpha_o(q,\theta)$ but now track its mean and variance across multiple trials:
\begin{equation}
   \mu_o^{\text{inc}}(\theta) =\sum_{q\in Q} \mathbb{E}[\alpha_o(q,\theta)] = \frac{1}{N_t} \sum_{n=1,q\in Q}^{N_t} \alpha_o^n(q,\theta)
   \label{eq:mean_inc}
\end{equation}
\begin{align}
   \sigma_o^{\text{inc}}(\theta)^2 &= \sum_{q\in Q} \text{Var}[\alpha_o(q,\theta)] \nonumber \\
   &= \frac{1}{N_t} \sum_{n=1}^{N_t} \sum_{q \in Q} \left( \alpha_o^n(q,\theta) - \mu_o^{\text{inc}}(\theta) \right)^2
   \label{eq:var_inc}
\end{align}

For exclusive randomization, we modify our framework to exclude the original position from the randomization explicitly set $\mathcal{O}_o' = \mathcal{O} \setminus \{o\}$, giving us similarly $\mu_o^{\text{exc}}(\theta)$ and $\sigma_o^{\text{exc}}(\theta)^2$.

For a given position $o$ and randomization parameter $\theta$, we select a $\theta$ portion of questions for randomization. In inclusive randomization, these questions can have their correct answer placed at any position, while in exclusive randomization, the correct answer is prevented from appearing at position $o$. The correct answer maintains its original position for the remaining $(1-\theta)$ portion of questions.

The difference between these metrics provides insight into how strongly the model associates specific positions with correct answers:
\begin{equation}
   \Delta\mu_o(\theta) = \mu_o^{\text{inc}}(\theta) - \mu_o^{\text{exc}}(\theta)
   \label{eq:delta_mean}
\end{equation}

Our analysis reveals striking position-dependent behaviors under both randomization protocols. Under inclusive randomization (\cref{fig:acc_inc}), positions exhibit markedly different initial accuracies, with position A showing highest accuracy ($\mu_A^{\text{inc}}(0) \approx 0.429$) and position D showing lowest ($\mu_D^{\text{inc}}(0) \approx 0.235$). As $\theta$ increases toward 1, these accuracies gradually converge to the uniform random chance level of 0.25, reflecting complete randomization of answer positions.

The exclusive randomization case (\cref{fig:acc_exc}) demonstrates intriguingly different dynamics. Most notably, the accuracy curves for all positions intersect at $\theta = 0.5$, suggesting a critical point where position-specific biases and randomization effects achieve perfect balance. Beyond this point, the positions again diverge before approaching their final randomized states.

\begin{figure}[!h]
   \centering
   \includegraphics[width=\linewidth]{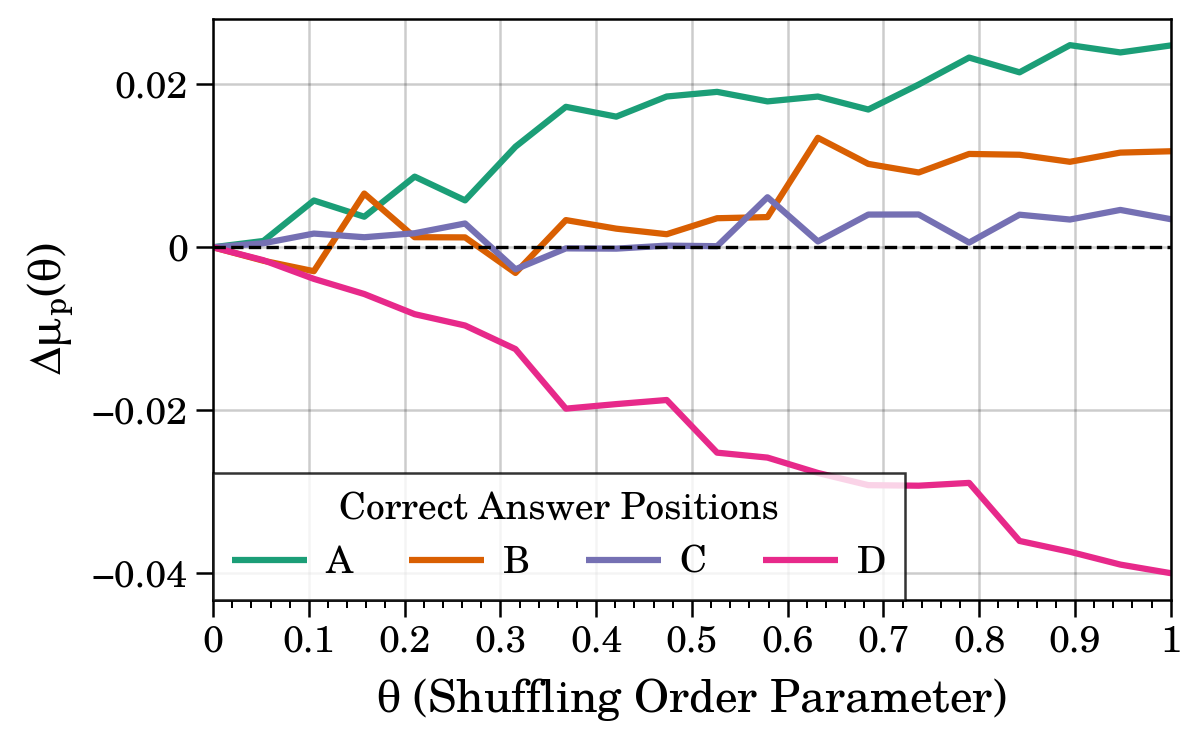}
   \caption{Difference $\Delta\mu_o(\theta)$ between inclusive and exclusive randomization accuracies. Position D shows strongest bias with monotonically decreasing difference reaching -0.04 at $\theta = 1$, while positions A, B, C maintain near-zero differences ($|\Delta\mu_o(\theta)| < 0.01$), suggesting more uniform treatment.}
   \label{fig:acc_diff}
\end{figure}

The difference metric $\Delta\mu_o(\theta)$ (\cref{fig:acc_diff}) reveals rich insights into how randomization affects position-specific biases. As $\theta$ increases from 0, positions A, B, and C show relatively small fluctuations around zero ($|\Delta\mu_o(\theta)| < 0.01$), suggesting these positions maintain similar accuracies whether or not they can receive their original correct answers during randomization. However, position A demonstrates a slight but consistent positive bias ($\Delta\mu_A(\theta) \approx 0.02$) for larger $\theta$, indicating it retains some advantage even under heavy randomization. In stark contrast, position D exhibits a strong monotonic decrease in $\Delta\mu_D(\theta)$, reaching -0.04 as $\theta \to 1$, revealing a systematic disadvantage when randomization prevents correct answers from appearing in this position. This pronounced negative bias, combined with positions B and C showing nearly identical neutral behavior, suggests the model has developed a strong aversion to position D that persists even under extensive randomization. The variance plots in both scenarios complement this finding with characteristic inverted U-shapes peaking at intermediate $\theta$ values, marking regions of maximum prediction uncertainty as the model transitions between position-dependent and randomized behavior modes.

\begin{figure}[!h]
\centering
\includegraphics[width=\linewidth]{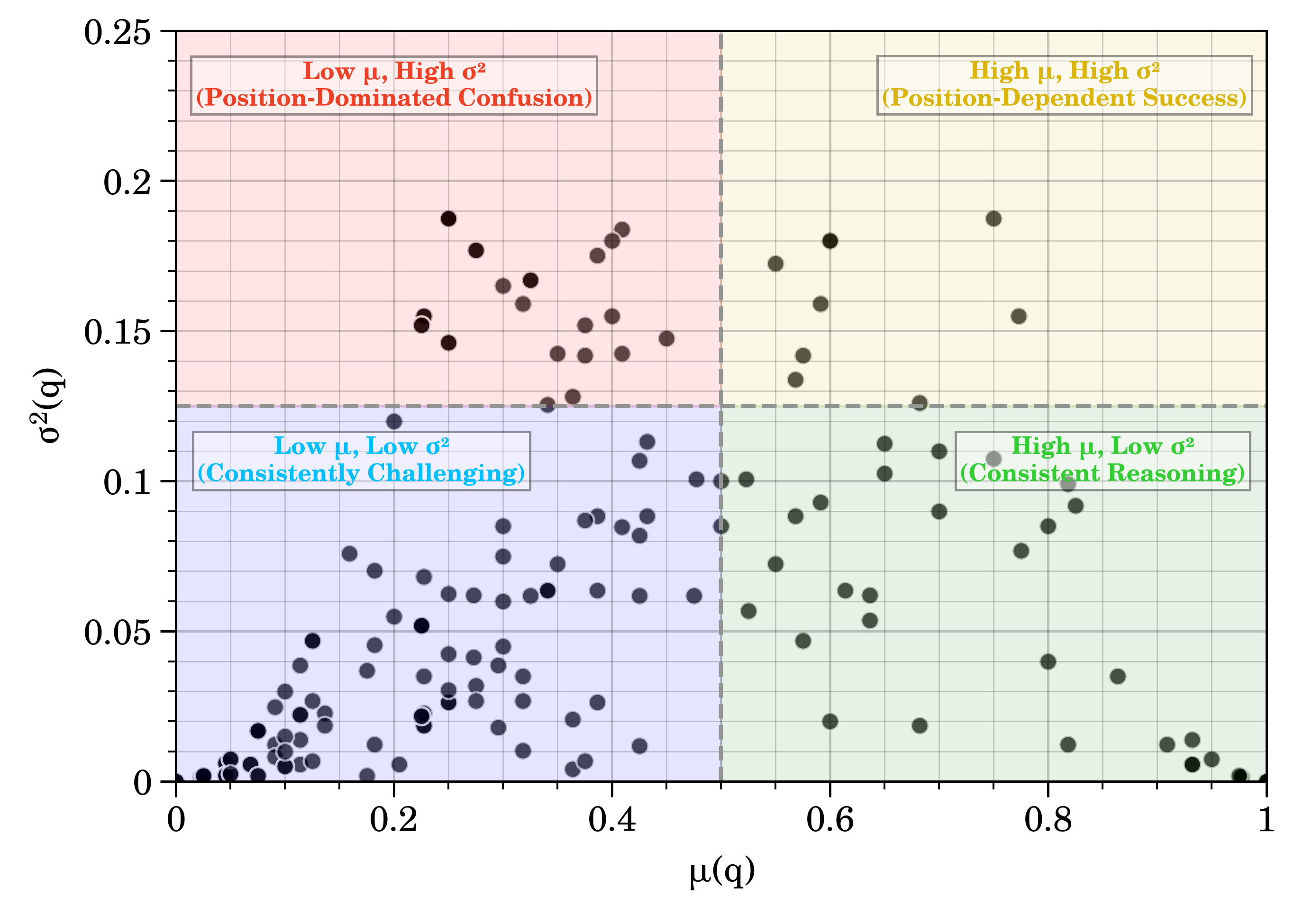}
\caption{Distribution of questions in the accuracy-variance space, revealing distinct patterns of model behavior. Questions cluster into four regions based on their position-averaged accuracy $\mu(q)$ and positional variance $\sigma^2(q)$ (\cref{eq:avg_acc,eq:pos_var}) in the two-dimensional difficulty map $\Psi(q,\theta)$ with $\theta$ integrated out. The space naturally partitions into regions reflecting distinct reasoning behaviors: consistent reasoning, position-dependent success, consistently challenging, and position-dominated confusion.}

\label{fig:phase_space}
\end{figure}

The position-dependent accuracy analysis reveals complex dynamics in model behavior, particularly evident in the systematic biases observed across different answer positions. To further elucidate these patterns, we examine the distribution of questions in the $\mu$-$\sigma^2$ phase space (\cref{fig:phase_space}). Drawing from our formalism in \cref{eq:avg_acc,eq:pos_var}, $\mu(q)$ captures the position-averaged accuracy across all positions, while $\sigma^2(q)$ quantifies the variance in position-dependent performance (where we have integrated out $\theta$ dependency). These metrics form a natural bounded space: $\mu(q) \in [0,1]$ follows directly from its definition as a mean probability, while $\sigma^2(q) \in [0,0.25]$ emerges from the fundamental properties of variance in binary outcomes with mean $\mu$. 

The phase space reveals four distinct regions of model behavior, each offering unique insights into the interplay between reasoning capability and positional dependence. Questions clustering in the high-$\mu$, low-$\sigma^2$ region ($\mu > 0.5$, $\sigma^2 < 0.125$) demonstrate the most substantial evidence of genuine reasoning—these are problems the model solves consistently regardless of answer position, suggesting reliance on logical deduction rather than positional cues. In striking contrast, the high-$\mu$, high-$\sigma^2$ region reveals questions where strong performance appears to be achieved through position-dependent strategies. These questions highlight how positional bias can inflate perceived reasoning capabilities. The model's success here likely stems from learned heuristics rather than substantive understanding, as evidenced by the significant variation in performance across different answer arrangements.

The low-$\mu$ regions provide complementary insights into model limitations. Questions in the low-$\mu$, low-$\sigma^2$ quadrant ($\approx 23\%$ of the dataset) present consistent challenges to the model while maintaining position independence, suggesting fundamental conceptual difficulties rather than failures of positional heuristics. Perhaps most revealing is the low-$\mu$, high-$\sigma^2$ region, where poor overall performance combines with high positional sensitivity—a clear indicator of reasoning breakdown and reliance on conflicting positional cues. This observation reinforces a central theme: \textit{accuracy alone often overstates a model's reasoning abilities}. A more refined approach is essential to uncover the complexities of underlying behavior, as explored in subsequent sections.

\subsection{Decomposing Language Model Decisions: Reasoning, Memorization, or Guessing?}\label{subsec:result-pmm}

Having established the prevalence of positional dependencies in model behavior, we now turn to a more fundamental question: what cognitive strategies drive these responses? Our probabilistic mixture model, introduced in \cref{eq:strategy_norm,eq:p_correct_mix}, provides a principled framework for decomposing model behavior into three fundamental strategies—pure logical reasoning (applying systematic deduction), pattern-based memorization (leveraging learned position-specific heuristics), and random guessing (selecting answers without clear rationale). This decomposition builds directly on our positional bias observations, where varying performance across positions suggested the interplay of multiple decision-making processes. By quantifying the relative contributions of each strategy, we move beyond aggregate performance metrics to understand the underlying mechanisms that shape model predictions.

The key insight of our analysis lies in leveraging positional information to distinguish between these strategies. As formalized in \cref{eq:p_mem}, memorization manifests as strong position-specific preferences, while \textit{true} reasoning should exhibit position-invariant performance. By measuring the accuracy differential between memorized and non-memorized positions (\cref{eq:delta_acc}), we can isolate the contribution of memorization ($P_M(q)$) and subsequently derive the probabilities of reasoning ($P_R(q)$) and guessing ($P_G(q)$) through \cref{eq:p_reason,eq:p_guess}. This approach capitalizes on the position-dependent effects observed in \cref{subsec:position analysis}, using systematic variations in performance across positions as a probe for understanding strategy selection. This analysis reveals striking patterns in strategy selection across different question types.

\begin{figure}[!htb]
\centering
\includegraphics[width=\linewidth]{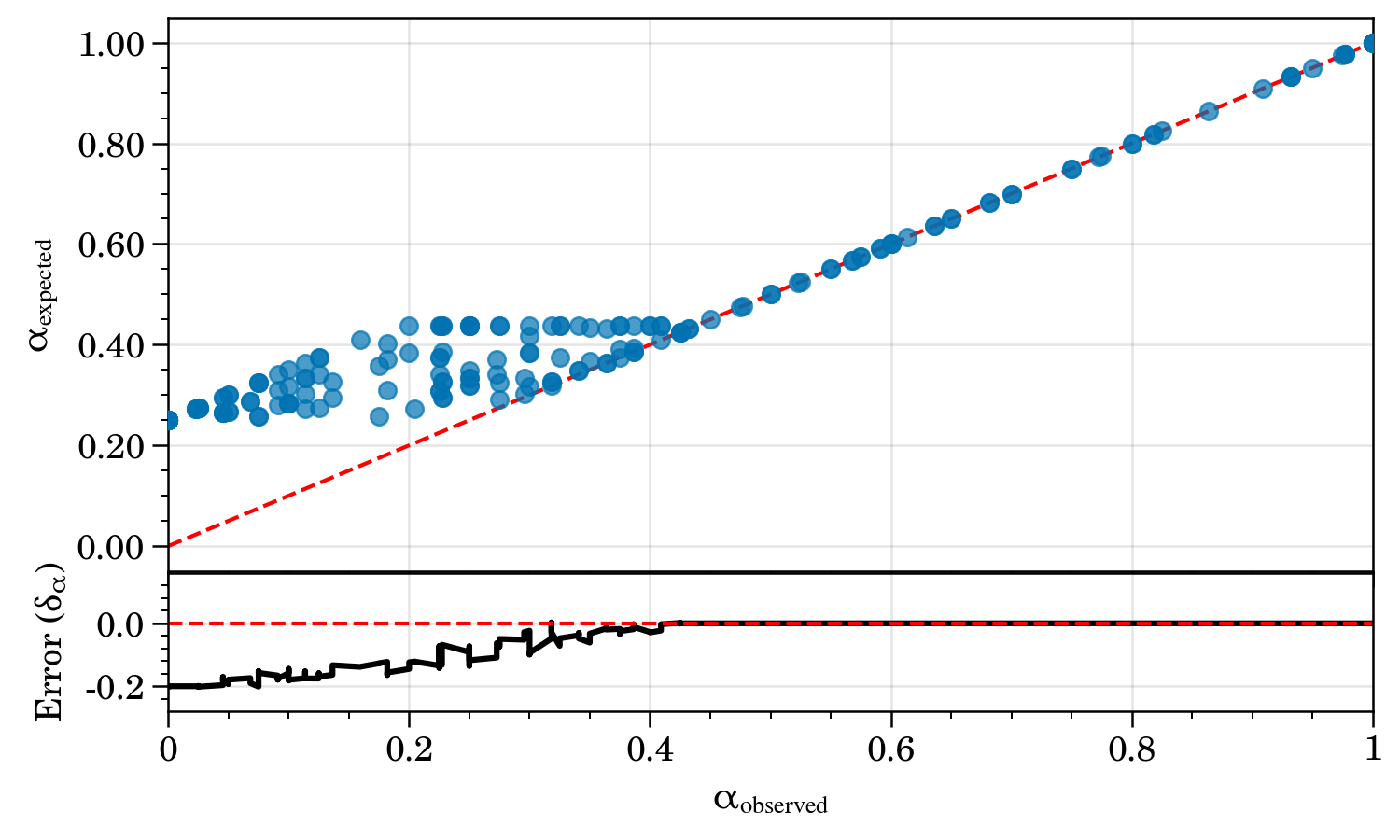}
\caption{Validation of the probabilistic mixture model. Comparison between model-predicted accuracies ($\alpha_{\text{expected}}$) and empirically observed accuracies ($\alpha_{\text{observed}}$) across all questions. The model demonstrates strong predictive power for medium to high-accuracy questions while exhibiting deviations in the low-accuracy regime.}
\label{fig:model_validation}
\end{figure}

Before examining the full distribution of cognitive strategies, we first establish the validity of our phenomenological model through rigorous comparison with empirical data. For each question $q$, we derive the position-averaged observed accuracy as:
\begin{equation}
\alpha_{\text{observed}}(q) = \frac{\alpha_{o_m}(q) + 3\alpha_{\text{other}}(q)}{4},
\end{equation}
where $\alpha_{o_m}(q)$ represents accuracy when the correct answer is at the memorized position, $\alpha_{\text{other}}(q)$ for other positions, and the factor of 3 accounts for the three non-memorized positions in our four-option setting. Similarly, the model's expected accuracy follows:
\begin{equation}
\alpha_{\text{expected}}(q) = \frac{\mathbb{E}\big[\alpha_{o_m}(q)\big] + 3\mathbb{E}\big[\alpha_{\text{other}}(q)\big]}{4},
\end{equation}
with expected accuracies derived from our strategy probabilities: $E[\alpha_{o_m}(q)] = P_M(q) + P_R(q) + P_G(q)P_{\mathcal{O}}$ and $E[\alpha_{\text{other}}(q)] = P_M(q)P_{\mathcal{O}} + P_R(q) + P_G(q)P_{\mathcal{O}}$ where $P_{\mathcal{O}}=1/4$ represents the baseline random chance in our four-option setting.

The model demonstrates near-perfect predictive accuracy across a wide range of questions, as evidenced by the close alignment between expected and observed accuracies shown in \cref{fig:model_validation}, along with deviations
\begin{equation}
    \delta_{\alpha} = \big|\alpha_{\text{observed}}(q) - \alpha_{\text{expected}}(q)\big|, \label{eq:residual_model_error}
\end{equation}
where $\delta_{\alpha}$ quantifies how well our theoretical predictions match empirical observation. Particularly for questions with $\alpha_{\text{observed}}(q) > 0.4$, the agreement validates our core assumptions of reasoning and uniform guessing. The systematic deviations that emerge in the low-accuracy regime ($\alpha_{\text{observed}}(q) < 0.4$) offer interesting insights into the limitations of our first-order approximations, where we assume strategies combine linearly (especially with $\alpha_{\text{expected}}\not< 0.2$). Several higher-order effects likely contribute to these deviations: position-dependent reasoning introduces cross-terms between strategy selection and positional bias that our linear decomposition neglects; non-uniform distractor selection preferences modify the effective guessing probability; and the assumption of perfect reasoning becomes increasingly approximate as partial understanding leads to probabilistic reasoning outcomes. These effects, most pronounced where positional and heuristic dependencies dominate, suggest natural extensions to our theory—incorporating higher-order correlations between positional bias and strategy selection may capture the full complexity of the model's behavior in this regime. 

\begin{figure}[!htb]
    \centering
    \includegraphics[width=\linewidth]{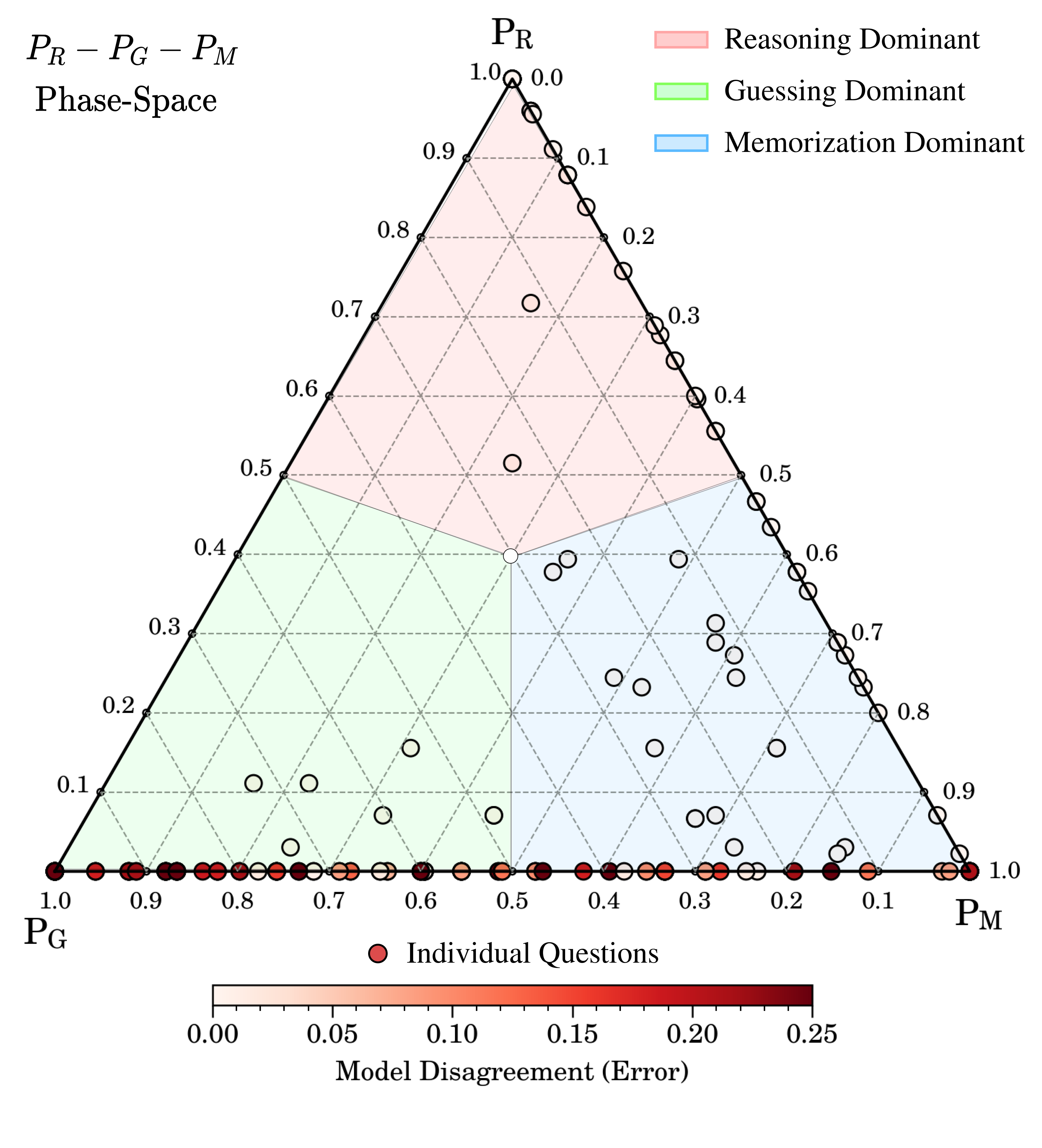}
    \caption{
Distribution of cognitive strategies across the probability simplex formed by Reasoning ($P_R$), Guessing ($P_G$), and Memorization ($P_M$) probabilities. Points represent individual questions with shading intensity corresponding to the model agreement (\cref{eq:residual_model_error}). The visualization reveals systematic clustering along edges rather than uniform coverage, indicating a preference for binary strategy combinations. The notable absence of points along the $P_R$-$P_G$ edge demonstrates persistent memorization contribution across all decision-making scenarios, while darker shading in memorization-dominated regions reflects increased model-observation deviation in these regimes (points/questions that do not fit well to our approximation, indicating contribution from higher order effects)
}
    \label{fig:ternary_plot}
\end{figure}

To further investigate these model behaviors and their limitations, we visualize the distribution of cognitive strategies through a ternary plot analysis (\cref{fig:ternary_plot}), where each point represents a question's unique mixture of reasoning ($P_R$), memorization ($P_M$), and guessing ($P_G$) probabilities derived from our framework (\cref{eq:p_reason,eq:p_guess}). The ternary representation is particularly powerful here as it naturally captures the constraint $P_R + P_M + P_G = 1$ while revealing clustering patterns that illuminate the model's decision-making preferences. One can see systematic relationships between cognitive strategies, yielding three fundamental insights about their distribution. First, points cluster predominantly along the edges rather than showing uniform coverage, suggesting the model typically relies on binary combinations of strategies rather than balanced mixtures of all three. Most notably, we observe a complete absence of points along the reasoning-guessing vertex ($P_R$-$P_G$), indicating that memorization ($P_M$) always contributes to the model's decision-making process, even in scenarios dominated by reasoning and guessing. This persistent reliance on learned patterns or heuristics aligns with our earlier positional bias analysis (\cref{subsec:position analysis}), where position-dependent effects were observed across all performance regimes.

\begin{figure*}[!htb]
    \centering
    \includegraphics[width=\linewidth]{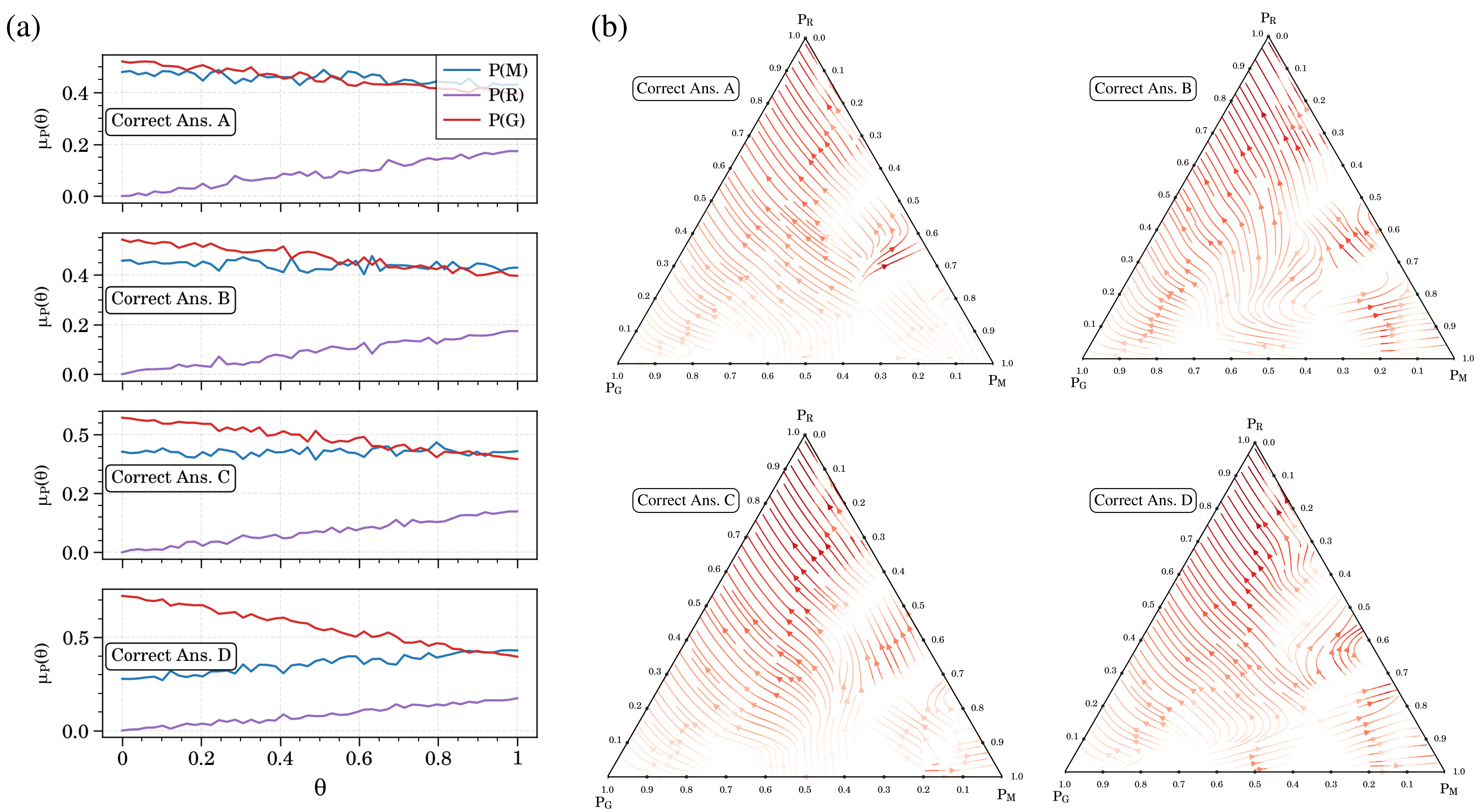}
\caption{
Evolution of cognitive strategy distributions and phase space behavior in language model decision-making. 
\textbf{(a)} Mean strategy probabilities $\mu_P^o(\theta)$ evolution across randomization parameter $\theta$ for each answer position $o \in \mathcal{O}$. The increase in reasoning contribution $\mu_R^o(\theta)$ with $\theta$ demonstrates transition to position-independent strategies under randomization.
\textbf{(b)} Probability flow dynamics in strategy space $\Delta^2$ showing position-dependent adaptation mechanisms. Positions $A$, $B$, $C$ exhibit coherent laminar flows with preserved memorization-reasoning ratios, while position $D$ shows complex vortical structures indicating strategy instability. Stable attractors emerge in moderate reasoning-memorization regions ($P_R \approx 0.4$-$0.5$, $P_M \approx 0.3$-$0.4$), suggesting optimal configurations between deductive reasoning and pattern recognition.
}
    \label{fig:phase-space-flows}
\end{figure*}

The presence of points in the interior of the triangle, though sparse, reveals cases where the model employs a more nuanced mixture of all three strategies. These interior points typically appear in regions where $P_M > P_R > P_G$, suggesting that when the model engages in significant reasoning, it tends to do so in conjunction with memorized patterns rather than in isolation. The shading in \cref{fig:ternary_plot} visualizes the model agreement ($\delta_{\alpha}$, \cref{eq:residual_model_error}) discussed earlier in \cref{fig:model_validation}, now decomposed across individual questions in the strategy probability space. Points showing the poorest agreement (darker shading) concentrate along the bottom edge where $P_R \approx 0$ with $P_M$ and $P_G$ being the contributing factors, precisely matching our earlier observations about systematic deviations in the low-accuracy regime. This concentration of model disagreement in regions dominated by memorization and guessing suggests that our linear decomposition is most accurate when reasoning plays a significant role.

While our static analysis of strategy distributions provides valuable insights into the model's decision-making process, examining how these strategies evolve under controlled randomization reveals deeper mechanistic properties of model behavior. By tracking the evolution of reasoning ($P_R$), memorization ($P_M$), and guessing ($P_G$) probabilities as functions of the order parameter $\theta$, we can observe how the model adjusts its cognitive strategies as positional cues become increasingly unreliable. This dynamic analysis extends our framework beyond snapshot distributions, illuminating the underlying mechanisms through which the model could transitions between different cognitive modes.

For a given correct answer position $o \in \mathcal{O}$, we calculate the mean strategy probabilities across all questions as:
\begin{equation}
   \mu_P^o(\theta) = \frac{1}{|\mathcal{Q}|} \sum_{q \in \mathcal{Q}} P_S(q, \theta, o)
   \label{eq:mean_strategy}
\end{equation}
where $S \in \{M, R, G\}$ represents the strategy (Memorization, Reasoning, or Guessing). This ensemble average quantifies the aggregate strategy preference for each position as randomization increases. These averages are computed using our previously derived quantities from \cref{eq:p_mem,eq:p_reason,eq:p_guess}, now extended to incorporate the randomization parameter $\theta$ and fixed correct answer position $o$. The position-specific accuracy differential under randomization becomes:
\begin{equation}
   \Delta \alpha_o(q, \theta) = \alpha_{o_m}(q, \theta, o) - \alpha_{\text{other}}(q, \theta, o)
   \label{eq:delta_acc_theta}
\end{equation}
which directly influences the decomposition of strategy probabilities through:
\begin{equation}
   P_M(q, \theta, o) = \frac{\Delta \alpha_o(q, \theta)}{1 - P_{\mathcal{O}}}
   \label{eq:p_mem_theta}
\end{equation}
where this relationship captures how positional preference evolves with increasing randomization.

Figure~\ref{fig:phase-space-flows}(a) visualizes these trajectories, revealing distinct characteristics in how the model adapts its decision-making approach as positional information becomes increasingly unreliable. The trajectories exhibit universal growth in reasoning contribution $\mu_R^o(\theta)$ across all positions $o \in \mathcal{O}$ as $\theta$ increases, a behavior that validates our theoretical framework—as positional cues become unreliable, the model necessarily shifts toward position-independent reasoning strategies. For $o \in \{A,B,C\}$, we observe similar baseline behaviors at $\theta = 0$, characterized by balanced but dominant distributions between memorization and guessing ($\mu_M^o \approx \mu_G^o \approx 0.4$) with minimal reasoning ($\mu_R^o < 0.1$). This initial equilibrium suggests these positions share common processing mechanisms in the model's architecture. This equilibrium largely persists through randomization, with gradual increases in the reasoning component.

Position $o = D$ exhibits markedly different characteristics that align with our positional bias analysis from \cref{subsec:position analysis}. At $\theta = 0$, it shows the highest guessing probability paired with minimal memorization ($\mu_M^D \approx 0.3$), indicating increased uncertainty. As $\theta$ increases, we observe a steady decrease in $\mu_G^D$ accompanied by an increase in $\mu_M^D$. This inverse relationship between guessing and memorization strategies at position $D$ suggests a fundamental difference in how the model processes information at sequence endings. For this language model and GPQA-like reasoning questions, these trajectories indicate that answers selected at position $D$ involve less memorization compared to other positions, though this comes with increased guessing rather than enhanced reasoning. This quantitative decomposition of strategy selection provides insights into positional bias effects that traditional performance metrics alone cannot capture.

Having established the position-dependent evolution of ensemble-averaged strategy probabilities, we now leverage our phenomenological framework to probe deeper patterns in the model's cognitive adaptation mechanisms. While the ensemble averages provide valuable insights, a more granular analysis of individual question trajectories with respect to our order parameter ($\theta$) in strategy space can validate our framework's predictions and potentially reveal emergent behavioral patterns.

For a question $q \in \mathcal{Q}$, its strategy distribution at $\theta$ is represented by the probability vector $\mathbf{P}_q(\theta) = (P_M(\theta), P_R(\theta), P_G(\theta))$ on the probability simplex $\Delta^2$ —a bounded, compact, and convex subset of $\mathbb{R}^3$ that naturally constrains our strategy probabilities to sum to unity. The evolution of these probabilities under increasing randomization defines a conservative dynamical system on $\Delta^2$:

\begin{equation}
   \frac{d\mathbf{P}_q}{d\theta} = \mathbf{F}_q(\mathbf{P}_q, \theta),
   \label{eq:strategy_flow_dynamics}
\end{equation}

where $\mathbf{F}_q$ represents the gradient field of strategy adaptation. Conservation of probability imposes fundamental constraints on this dynamics:

\begin{equation}
   \nabla \cdot \mathbf{F}_q = 0, \quad \sum_{i \in \{M,R,G\}} P_i(\theta) = 1.
   \label{eq:probability_constraints}
\end{equation}

\Cref{fig:phase-space-flows}(b) visualizes these probability flows through streamlines in the ternary space. Given the finite sampling of GPQA (198 questions), we employ symplectic interpolation \cite{wang2021key} across $\Delta^2$ at each $\theta$ to construct a continuous gradient field that preserves the conservative nature of the probability flow. The resulting stream plots reveal how questions traverse strategy space under increasing randomization, with distinct flow patterns emerging for different fixed correct answer positions $p \in \mathcal{O}$.

The flow patterns emerging from our dynamical analysis (\cref{eq:strategy_flow_dynamics,eq:probability_constraints}) provide quantitative insights into the cognitive mechanisms underlying language model behavior. For positions $o \in \{A,B,C\}$, we observe remarkably coherent laminar flows in \cref{fig:phase-space-flows}(b) that maintain consistent ratios between memorization and reasoning components even under increased randomization. This stratification suggests the existence of conserved quantities in strategy evolution, potentially reflecting fundamental constraints in how models balance different cognitive approaches.  In contrast, position $D$ exhibits complex vortical structures and multiple critical points, where the flow field's topology indicates regions of strategy space in which small perturbations to input structure can lead to significant shifts in model behavior. This mathematical characterization helps explain previously observed but poorly understood phenomena in language model responses, such as performance degradation when key information appears at sequence endings.

The emergence of stable attractors in moderate reasoning-memorization regions ($P_R \approx 0.4$-$0.5$, $P_M \approx 0.3$-$0.4$) has particular significance for understanding model capabilities and limitations. These points represent robust configurations where model behavior remains consistent despite positional perturbations, suggesting natural \textit{``sweet spots''} in the trade-off between pattern matching and logical deduction. This observation provides mathematical grounding for empirically successful prompting techniques like Chain-of-Thought (CoT)  \cite{wei2022chain}, where the explicit positional structure (e.g., \texttt{Step 1:}, \texttt{Step 2:}) acts as an external field that stabilizes the model's strategy selection. Consider the empirically observed phenomenon in letter counting tasks—specifically with the word ``strawberry''—where models consistently converge to a stable strategy configuration despite diverse reasoning attempts. The introduction of structured decomposition prompts (e.g., \textit{think step by step})  destabilizes this attractor state, enabling a transition to regions of higher reasoning probability in the strategy phase space. This phase transition in cognitive strategy demonstrates how positional structure in prompts can maintain balanced ratios of memorization and reasoning components. The existence of these stable attractors raises an intriguing question: Do these represent fundamental computational constraints of transformer architectures, or are they artifacts of current training methodologies?

Similarly, the effectiveness of instruction tuning can be understood through this lens—the consistent positioning of task specifications and completion patterns establishes reliable attractors in strategy space. Vortices, particularly prominent in position D, indicate configurations where the model cycles through different strategy combinations rather than converging to stable behavior. These structures suggest that specific prompt patterns may inherently lead to unstable responses, not due to limitations in model capability but rather as a natural consequence of how position-dependent information processing interacts with strategy selection. These structures suggest that specific prompt patterns may inherently lead to unstable responses, not due to limitations in model capability but rather as a natural consequence of how position-dependent information processing interacts with strategy selection.

\subsection{Information-Theoretic Frontier of Model Confidence}
\begin{figure}[!ht]
    \centering
    \includegraphics[width=\linewidth]{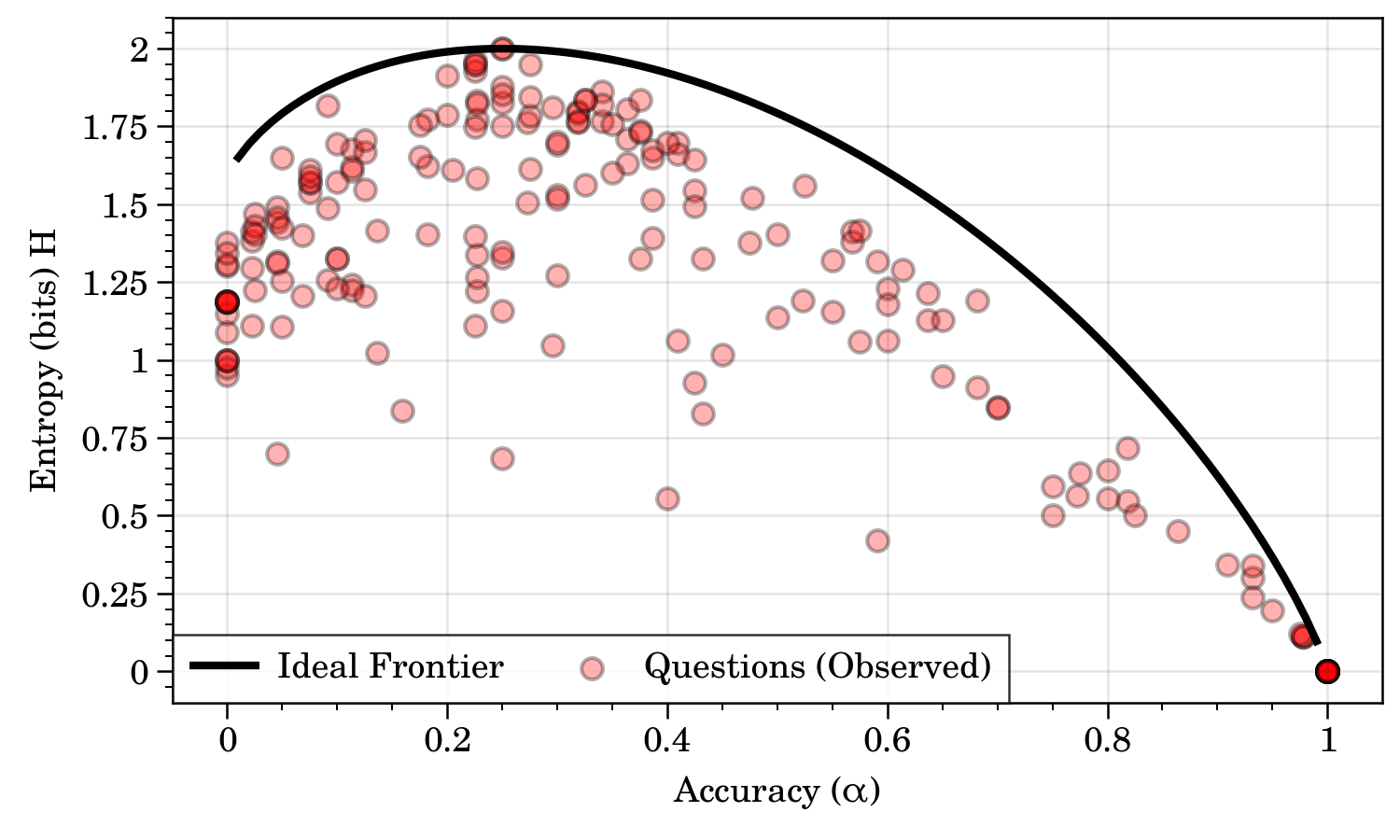}
    \caption{Distribution of questions in the entropy-accuracy space relative to the theoretical calibration frontier. Each point represents a question with empirical accuracy $\alpha$ and entropy $H$. The solid black curve shows the theoretical maximum entropy achievable at each accuracy level for a four-option multiple-choice system. All observed questions fall below this frontier, indicating systematic under-dispersion of probability mass compared to ideal calibration.}
    \label{fig:entropy_frontier_empirical}
\end{figure}

While our probabilistic mixture model effectively characterizes the fundamental strategies underpinning model decisions, it does not directly quantify the relationship between model confidence and strategy selection. To systematically investigate this relationship, we adopt the information-theoretic framework outlined in \cref{eq:entropy,eq:ideal_entropy}. This framework enables the analysis of prediction uncertainty in relation to both accuracy and the distribution of cognitive strategies. The theoretical entropy-accuracy frontier, as defined in \cref{eq:entropy_frontier}, serves as a benchmark for optimal model calibration. It represents the ideal probability mass distribution between correct and incorrect predictions across varying accuracy levels. We quantitatively assess the model's strategy selection and its associated uncertainty by examining deviations between empirical entropy-accuracy pairs and this theoretical frontier. This approach uncovers systematic variations in model uncertainty across different cognitive modes, providing mechanistic insights into how strategy selection influences prediction confidence. Integrating these entropy-based measurements with our probabilistic decomposition framework facilitates a comprehensive characterization of model behavior within the strategy probability simplex. This integration elucidates the mathematical relationships between accuracy, uncertainty, and cognitive strategy selection, offering a nuanced understanding of the interplay between these factors.
\begin{figure*}[!htb]
   \centering
   \includegraphics[width=\linewidth]{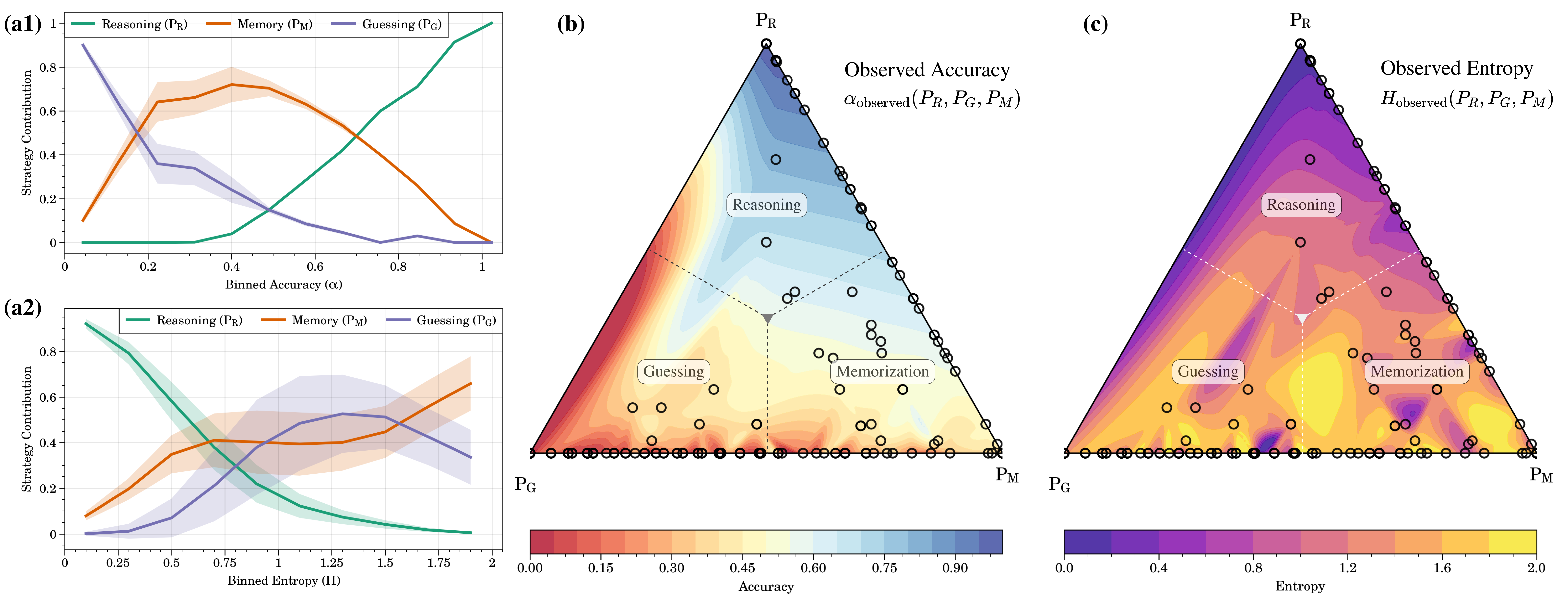}
   \caption{Distribution of cognitive strategies across accuracy and entropy spaces. \textbf{($a_1$)} Strategy contributions as a function of binned accuracy $\alpha$, revealing transitions from guessing to reasoning-dominated behavior with memorization peaking in intermediate regimes. The shaded regions represent one standard deviation in strategy probabilities within each bin. \textbf{($a_2$)} Strategy contributions versus binned entropy $H$, demonstrating systematic relationships between prediction uncertainty and strategy selection. \textbf{(b)} Observed empirical accuracy distribution in probability simplex. The scalar field shows interpolated accuracy values across strategy probability space, with dots representing individual questions. High accuracy (blue) concentrates near pure reasoning ($P_R$), while low accuracy (red) dominates near pure guessing ($P_G$). A moderate accuracy plateau (yellow) emerges in memorization-dominated regions. \textbf{(c)} Observed entropy distribution across the probability simplex. The scalar field represents interpolated entropy values, revealing complex topology with multiple local extrema. Low entropy (purple) concentrates near pure reasoning, while higher entropy (yellow) emerges in mixed strategy regions.
}
   \label{fig:strategy_distributions}
\end{figure*}
Examining the distribution of questions in entropy-accuracy space reveals fundamental deviations from the ideal probabilistic system $\mathcal{B}^*$ defined in \cref{eq:ideal_dist}. \Cref{fig:entropy_frontier_empirical} demonstrates that empirical entropy-accuracy pairs systematically fall below the theoretical calibration frontier derived in \cref{eq:entropy_frontier}, indicating a universal under-dispersion of probability mass across all accuracy levels. This systematic deviation from optimal probability distributions reveals that the model’s internal representations exhibit \textit{unwarranted determinism relative to their predictive performance}. The deviation pattern's magnitude varies non-uniformly with accuracy, providing insights into the model's confidence calibration mechanisms. In the low to moderate accuracy regime ($\alpha < 0.7$), questions exhibit entropy values substantially below $H_{\text{ideal}}(\alpha)$, with typical deviations of 0.2-0.5 bits.

This persistent entropy gap quantifies a fundamental limitation in the model's ability to express appropriate uncertainty when faced with challenging questions. The gap narrows notably for high-accuracy questions ($\alpha > 0.7$), suggesting that confidence scaling becomes more appropriate only when the model's internal evidence strongly supports a particular answer. This systematic overconfidence pattern provides a potential mechanistic explanation for the positional biases observed in \cref{subsec:position analysis}. The model's tendency to produce overly concentrated probability distributions may arise from excessive reliance on learned positional heuristics, leading to artificially inflated confidence even when such heuristics provide limited predictive value.

\subsection{Geometric Characterization of Strategy-Uncertainty Interplay}

To quantify how strategy selection relates to uncertainty calibration, we analyze the distribution of probabilistic mixture components across both accuracy and entropy spaces. \Cref{fig:strategy_distributions} demonstrates how reasoning ($P_R$), memorization ($P_M$), and guessing ($P_G$) probabilities vary systematically with predicted accuracy and uncertainty.

The accuracy-dependent strategy distribution (\cref{fig:strategy_distributions}.$a_1$) reveals distinct cognitive regimes as predictive performance varies. In low accuracy regions ($\alpha < 0.2$), the apparent dominance of guessing ($P_G \approx 0.9$) coincides with elevated model fit residuals (\cref{eq:residual_model_error}), suggesting our decomposition may not fully capture the complexity of model behavior in this regime. As accuracy increases ($0.2 < \alpha < 0.6$), memorization becomes the primary strategy ($P_M \approx 0.7$) while reasoning probability gradually rises. For high accuracy predictions ($\alpha > 0.6$), pure reasoning dominates ($P_R > 0.6$) with memorization and guessing diminishing significantly. This distribution of strategies aligns with theoretical expectations for an ideal reasoning system. \textit{High accuracy outcomes fundamentally require systematic deductive processes}—achieving consistent, correct predictions through memorization or guessing alone is statistically improbable. The observed dominance of $P_R$ in high-accuracy regions validates this theoretical constraint while revealing the transitional regimes where multiple strategies interact. The entropy-strategy relationship (\cref{fig:strategy_distributions}.$a_1$) provides complementary validation of our framework. Low entropy predictions ($H < 0.5$) exhibit high reasoning probabilities ($P_R \approx 0.8$), consistent with the theoretical requirement that confident correct answers must emerge from systematic deduction rather than arbitrary selection. The transition point at $H \approx 0.75$, where all strategies roughly contribute equally, marks a critical uncertainty threshold above which the model shifts from reasoning to increasingly heuristic approaches.

This quantitative decomposition of strategy selection provides a novel framework for evaluating language model capabilities. Our dual analysis through PPM and ITC frameworks reveals fundamental relationships between accuracy, uncertainty, and cognitive strategies that can inform the principled deployment of these models. The dominance of reasoning ($P_R$) in high-accuracy, low-entropy regimes validates that genuine logical deduction, rather than memorization or guessing, drives reliable model performance. This observation has significant implications for establishing quantitative thresholds in applications requiring verifiable reasoning. For instance, in domains where incorrect predictions carry significant consequences, one might require both high reasoning probability ($P_R > 0.7$) and low entropy ($H < 0.5$ bits), ensuring decisions emerge from systematic deduction rather than uncertain pattern matching\footnote{These curves are likely specific to the distribution of the dataset and the characteristics of the model. In our case, the questions predominantly involve queries that are heavily reliant on \textit{reasoning}.}. The persistent contribution of memorization ($P_M$), even in reasoning-dominated regimes, suggests that complete elimination of learned heuristics may be neither possible nor desirable. Instead, applications should specify acceptable ratios between strategies based on their requirements. A scientific question-answering system might tolerate moderate memorization ($P_M < 0.3$) if accompanied by dominant reasoning ($P_R > 0.6$), while a medical diagnostic tool would demand stricter thresholds favoring pure reasoning.

Our entropy-strategy analysis provides additional validation criteria through uncertainty calibration. Applications can leverage the relationship between entropy and strategy distribution to identify cases where model confidence aligns with genuine reasoning capability. The critical threshold at $H \approx 0.75$ bits, where strategies achieve equal contribution, marks a natural boundary between reliable reasoning and heuristic-dominated predictions. This threshold, combined with accuracy-based criteria, enables fine-grained control over model deployment. For example, educational applications might accept predictions where $H < 1$ bit and $P_R + P_M > 0.8$, allowing some reliance on memorized patterns while maintaining reasonable confidence in the underlying comprehension.

Next, we examine how accuracy and entropy manifest as scalar fields across the probability space. The probability simplex $\Delta^2$ supports two fundamental scalar fields - accuracy $\mathcal{A}$ and entropy $\mathcal{H}$, which characterize the model's behavior at each strategy configuration:

\begin{equation}
   \mathcal{A}: \Delta^2 \rightarrow [0,1], \quad \mathcal{H}: \Delta^2 \rightarrow [0,2]
   \label{eq:field_mappings}
\end{equation}
\begin{table*}[!htp]
\centering

\begin{tabular}{lccc}
\toprule
\textbf{Metric} & Memorization$(P_M)$ & Reasoning$(P_R)$ & Guessing$(P_G)$ \\
\midrule
Accuracy & -0.006 & 0.912 & -0.767 \\
Entropy & 0.436 & -0.849 & 0.329 \\
\bottomrule\\
\end{tabular}
\caption{Correlation analysis between performance metrics and strategy probabilities. Values show Pearson correlation coefficients with all correlations being statistically significant ($p < 10^{-3}$) except for $P_M$ vs Accuracy ($p \approx 0.9$).}
\label{tab:correlations}
\end{table*}

These fields emerge naturally from our question set $\mathcal{Q}$, where each question $q$ provides a measurement $(\alpha_q, H_q)$ at its corresponding strategy point $\mathbf{P}_q$. Similar to the strategy established in \cref{eq:strategy_flow_dynamics}, these discrete measurements can be smoothly interpolated to reveal the complete field structure. Analysis of these interpolated surfaces provides insights into how strategy selection influences model performance and uncertainty across the full space of cognitive approaches.

The interpolated accuracy field reveals a rich structure in strategy-dependent model performance (\cref{fig:strategy_distributions}.b). Peak accuracy ($\alpha > 0.9$) concentrates near the $P_R$ vertex, quantitatively validating that high-performance prediction requires substantial reasoning contribution. This aligns with our theoretical framework where genuine logical deduction, rather than pattern matching or random selection, enables reliable problem-solving. A continuous gradient emerges from the $P_G$ vertex toward higher $P_R$ regions, demonstrating systematic improvement in performance as strategies transition from guessing to reasoning-dominated approaches. The field structure exhibits a notable plateau of moderate accuracy ($0.45 < \alpha < 0.6$) in regions where memorization dominates ($P_M > 0.6$). This characteristic suggests memorization can achieve limited but consistent performance, potentially explaining the stable position-dependent behaviors observed in our randomization analysis. The sharp accuracy gradient near the $P_R$ vertex corresponds to the entropy-accuracy frontier results from \cref{fig:entropy_frontier_empirical}, where high accuracy correlates with concentrated probability distributions. 

The complementary entropy field (\cref{fig:strategy_distributions}.c) reveals fundamental characteristics of the model's uncertainty calibration across different cognitive modes. The concentration of minimum entropy near the $P_R$ vertex demonstrates that reasoning-dominated predictions exhibit high confidence, aligning with theoretical expectations for systematic deduction. This topology supports our analysis from \cref{eq:entropy_frontier}, where deviations from ideal calibration were smallest when reasoning dominated the decision process. A distinctive entropy valley emerges along the $P_R$-$P_M$ edge, indicating that both pure strategies can produce confident predictions through different mechanisms. However, the entropy structure becomes markedly more complex in regions of mixed strategy contributions, manifesting as multiple local maxima. These features suggest that uncertainty increases when the model actively negotiates between competing cognitive approaches. The emergence of high-entropy pockets in memorization-dominated regions provides a mechanistic explanation for the position-dependent confidence variations observed in our randomization analysis. This pattern indicates that memorization-based predictions inherently carry more significant uncertainty, particularly when positional cues provide conflicting evidence.

Furthermore, the non-uniform entropy distribution across mixed-strategy regions unveils complex interactions between cognitive modes that transcend simple linear combinations. These patterns, aligned with the flow structures identified in \cref{eq:strategy_flow_dynamics}, suggest that entropy variations serve as underlying drivers of transitions between cognitive regimes. The topological complexity of this landscape indicates that prediction confidence emerges from both the dominant strategy and the specific configuration of cognitive approaches employed.

Complementing \cref{fig:strategy_distributions}(a), analysis of the quantitative relationships between cognitive strategies and model performance metrics reinforces our theoretical framework through statistical validation. Correlation analysis (\cref{tab:correlations}) reveals a rich relationship structure between strategy probabilities, accuracy, and entropy. The strong positive correlation between reasoning probability and accuracy ($\rho_{\alpha,P_R} = 0.912$) provides compelling evidence that reliable performance emerges from systematic deductive processes. This finding gains additional support from the complementary negative correlation with guessing ($\rho_{\alpha,P_G} = -0.767$), mathematically demonstrating how random selection fundamentally opposes consistent performance. The entropy analysis adds another critical dimension - reasoning shows a strong negative correlation with entropy, indicating that systematic deduction leads to more concentrated probability distributions and higher confidence predictions. Conversely, memorization exhibits a moderate positive correlation with entropy, suggesting that reliance on learned patterns introduces inherent uncertainty in the prediction process. These relationships, emerging from our dual analysis framework, demonstrate how memorization and reasoning represent fundamentally different cognitive modes - with reasoning enabling higher accuracy and lower uncertainty in model predictions, as expected.

\section{Conclusion}

This work introduces a fundamentally new approach to understanding and evaluating reasoning capabilities in language models through the lens of phenomenological analysis. By developing dual frameworks - the Probabilistic Mixture Model (PMM) and Information-Theoretic Consistency (ITC) Analysis - we demonstrate how controlled experimentation (like positional bias) can reveal deeper insights into model behavior than traditional accuracy metrics alone. Our analysis reveals several critical findings about current language models' reasoning capabilities. First, the persistent coupling between positional structures and strategy selection, evidenced by non-zero memorization components even in reasoning-dominated regions, suggests that truly position-invariant logical deduction remains challenging for current models. The emergence of stable attractors in strategy space and the identification of conserved flow structures points to fundamental constraints in how these models balance different cognitive approaches - patterns that likely reflect architectural/training limitations rather than task-specific phenomena.

These results highlight the limitations of traditional benchmarks in evaluating language models. Aggregate accuracy metrics tend to overstate models’ reasoning capabilities by failing to account for critical factors such as memorization, random guessing, position-dependent effects, and the nuanced interplay of mixed strategies. Our framework demonstrates that genuine reasoning, characterized by high reasoning probability ($P_R$) and low entropy ($H$), emerges only under specific conditions. This regime occupies a relatively small portion of the strategy space, while the majority of apparent success relies on sophisticated combinations of memorization and pattern matching. This insight enables quantitative criteria for real-world deployments. For example, educational systems may tolerate moderate levels of memorization $(P_M < 0.3)$, while medical applications could demand strict reasoning thresholds, such as $(P_R > 0.7, H < 0.5~\text{bits})$, to ensure reliability.

Looking forward, this work opens several promising directions for future research. This work suggests the existence of underlying mathematical structures governing model responses - patterns that merit deeper theoretical investigation. While the field continues its essential work of mitigating systematic biases in language models, our framework demonstrates how these same biases can be harnessed as analytical tools, providing unique windows into model behavior and capabilities. By expanding this approach to examine other forms of systematic bias beyond positional effects, we can develop increasingly sophisticated methods for understanding how models combine different cognitive strategies and beyond. This deeper understanding of strategy interplay offers a path toward creating more nuanced benchmarks and evaluation methods, ultimately advancing our pursuit of \textit{genuine logical deduction in artificial intelligence systems}.

\bibliography{references}
\end{document}